\documentclass[a4paper,final]{IEEEtran}    
\usepackage{multirow}
\usepackage{amsfonts}
\usepackage{amssymb}
\usepackage{amsbsy}
\usepackage{amsmath}
\usepackage{array}
\usepackage{cite}
\usepackage{color}
\usepackage{booktabs}
\usepackage{lscape}
\usepackage{arydshln}
\usepackage[dvips]{graphicx}
\usepackage{url}
\usepackage{epsfig}
\usepackage{epstopdf}
\usepackage{subfigure}
\usepackage{algorithm}
\usepackage{algpseudocode}
\usepackage{hyperref}

\newcommand{\red}[1]{\textcolor [rgb]{1,0,0}{#1}}
\newcommand{\green}[1]{\textcolor [rgb]{0,0.6,0}{#1}}
\newcommand{\blue}[1]{\textcolor [rgb]{0,0,0.8}{#1}}

\newcommand{\RR}[2]{\textcolor [rgb]{0,0,0}{#2}}

\newcommand{\x}{\mathbf{x}}
\newcommand{\W}{\mathbf{W}}
\newcommand{\w}{\mathbf{w}}
\renewcommand{\v}{\mathbf{v}}

\newcommand{\balpha}{\boldsymbol\alpha}
\newcommand{\y}{\mathbf{y}}

\newcommand{\K}{\mathbf{K}}

\newcommand{\D}{\mathbf{D}}

 \def\R{{\mathbb{R}}} 

\newcommand{\mytitle}{Non-convex regularization in
  remote sensing}
\begin{document}

\title{\mytitle}
\author{Devis Tuia~\IEEEmembership{Senior Member,~IEEE}, Remi Flamary, Michel Barlaud,
\thanks{Manuscript received September x, 2014;} \thanks{This research was partially funded by the Swiss National Science Foundation, under grant PP00P2-150593.}
\thanks{\noindent DT is with the University of Zurich,
  Switzerland. Email: devis.tuia@geo.uzh.ch, web: http://geo.uzh.ch/ ,
  Phone: +4144 635 51 11, Fax: +4144 635 68 48.}
\thanks{\noindent MB is with is with the University of Nice Sophia
  Antipolis, France. }
\thanks{\noindent RF is with the University of Nice Sophia Antipolis,
  OCA and Lagrange Lab., France. Email: remi.flamary@unice.fr, web: remi.flamary.com}
\thanks{\noindent DT and  RF contributed equally to the paper.}
\thanks{\noindent Digital Object Identifier xxxx}}

\markboth{IEEE Trans. Geosci. Remote Sens., Vol. XX, No. Y, Month Z 2016}{Tuia and Flamary: \mytitle}

\maketitle

\begin{abstract}

In this paper, we study the effect of different regularizers and their
implications in high dimensional image
classification and {sparse linear unmixing}. Although kernelization or sparse methods are
globally accepted solutions for processing data in high dimensions, we present here a study on the impact of the form of
regularization used and its parametrization. We consider
regularization via traditional squared ($\ell_2$) and
sparsity-promoting ($\ell_1$) norms, as well as more unconventional
nonconvex regularizers ($\ell_p$ and Log Sum Penalty). We compare their properties and advantages on several
 classification and {linear unmixing} tasks and provide advices on the choice
of the best regularizer for the problem at hand. Finally, we also
provide a fully functional toolbox for the community\footnote{  \url{https://github.com/rflamary/nonconvex-optimization}}.

\end{abstract}

\section{Introduction}
\label{sec:intro}

Remote sensing image processing~\cite{Cam11} is a fast moving area of science. Data acquired from satellite or airborne sensors  and converted into useful information (land cover maps, target maps, mineral compositions, biophysical parameters) have nowadays entered many applicative fields: efficient and effective methods for such conversion are therefore needed. This is particularly true for data sources such as hyperspectral \RR{}{and very high resolution images, whose data volume is big and structure is complex: for this reason many traditional methods perform poorly when confronted to this type of data. The problem is even more exhacerbated when dealing with multi-source and mult-imodal data, representing different views of the land being studied (different frequencies, different seasons, angles, ...).} This created the need for more advanced techniques, often based on statistical learning~\cite{CampsValls14spm}.

Among such methodologies, regularized methods are certainly the most successful. Using a regularizer imposes some constraints on the class of functions to be preferred during the optimization of the model and can thus be beneficial if we know what  these properties are. The more often, regularizers are used to favour simpler functions over very complex ones, in order to avoid overfitting of the training data: in classification, the support vector machine uses this form of regularization~\cite{Cam05,Mou11}, while in regression examples can be found in kernel ridge regression or Gaussian processes~\cite{Ver12}. 

But smoothness-promoting regularizers are not the only ones that can
be used: depending on the properties one wants to promote, other
choices are becoming more and more popular. A first success story is
the use of Laplacian regularization~\cite{Bel05}: by enforcing
smoothness in the local structure of the data, one can promote the
fact that points that are similar in the input space must have a
similar decision function (Laplacian SVM~\cite{Gom07,Ben15} and
dictionary-based methods~ \cite{Zha15,Sun15}) or be projected close
after a feature extraction step (Laplacian eigenmaps~ \cite{Tu12} and
manifold alignment~\cite{Tui13d}). Another popular property to be
enforced, on which we will focus the rest of this paper, is
sparsity~ \cite{Don06}. Sparse models have only a part of the initial
coefficients which is active (i.e. non-zero) and are thus
compact. This is desirable in classification when the dimensionality
of the data is very high (e.g. when adding many spatial
filters~\cite{Fauvel13,tuia2014automatic} or \RR{}{using convolutional neural networks~ \cite{romero2016tgrs,DFCA}}) or in sparse coding when we
need to find a relevant \RR{}{dictionary} to express the
data~\cite{Ior11}. \RR{Recently}{Even though non-sparse models can work well in terms of overall accuracy, they still store information about the training samples to be used at test time: if such information is very high dimensional and the number of training samples is important, the memory requirements, the model complexity and -- as a consequence -- the execution time are strongly affected. Therefore, when processing next generation, large data using models generating millions of features~\cite{tokarczyk2014tgrs,Vol16b}, sparsity is very much needed to make models portable, while remaining accurate. For this reason}, sparsity has been extensively used in i)
spectral unmixing~ \cite{Bio12}, where a large variety of algorithms
is deployed to select  endmembers as a small fraction of the existing
data~ \cite{Ior11,Qu14,Ior14}, ii) image classification, where
sparsity is promoted to have portable models either at the level of
the samples used in reconstruction-based methods~\cite{Che11,Tan15} or
in feature selection schemes~\cite{Son14,tuia2014automatic,Tui15} iii)
and in more focused applications such as 3-D reconstruction from
SAR~\cite{Zhu12}, phase estimation~ \cite{Hon15} or
pansharpening~ \cite{Li10}.\\

{A popular approach to recover sparse features is to solve a convex optimization problem involving 
the $\ell_1$ norm (or Lasso) regularization~\cite{dsUP,doel,candes}. 
Proximal splitting methods have been shown to be highly effective
in solving sparsity-constrained problems~\cite{Smms05,Silvia,Siop07}.
The Lasso formulation based on the
penalty on the $\ell_1$ norm  of the model has been shown to be an
efficient shrinkage and sparse
model selection method in regression \cite{tRS,hrtzER,dlSR}.}
However, the Lasso regularizer is known to promote biased estimators
leading to suboptimal classification performances when strong sparsity
is promoted~\cite{candes2008enhancing,zhang2010analysis}. A way out of
this dilemma between sparsity and performance is to re-train a
classifier, this time non-sparse, after the feature selection has been
performed with Lasso~ \cite{tuia2014automatic}. Such scheme works, but
at the price of training a second model, thus leading to extra
computational effort and to the risk of suboptimal solutions, since we
are training a model with the features that were considered optimal by
another. In unmixing, synthetic ex\RR{e}{a}mples also show that the Lasso
regularization is not the one leading to the best abundance
estimation~\cite{Sig14}.

In recent years, there has been a trend in the study of unbiased sparse regularizers. These regularizers, typically the $\ell_0$, $\ell_q$ and Log Sum Penalty (LSP~\cite{candes2008enhancing}), can solve the dilemma between sparsity and performance, but are non-convex and therefore cannot be solved by known off-the-shelf convex optimization tools. Therefore, such regularizers have until now received little attention in the remote sensing community. A handful of papers using $\ell_q$ norm are found in the field of spectral unmixing~\cite{Qia11,Sig14,Wan15} where authors consider nonnegatrive matrix factorization solutions; in the modelling of electromagnetic induction responses, where the model parameters were estimated by regularized least squares estimation~\cite{Wei11}; in feature extraction using deconvolutional networks~\cite{Zha14} and in structured prediction, where authors use a non-convex sparse classifier to provide posterior probabilities to be used in a graph cut model~\cite{Jia15}. In all these studies, the non-convex regularizer outperformed the Lasso, while still providing sparse solutions.

In this paper, we give a critical explanation and  theoretical
motivations for the success of  regularized classification, with a
focus on non-convex methods. By comparing it with other traditional
regularizers (ridge $\ell_2$ and Lasso $\ell_1$), we advocate the use
of non-convex regularization in remote sensing image processing tasks:
non-convex optimization marries the advantages of accuracy and
sparsity in a single model, without the need of unbiasing in two steps
or reduce the level of sparsity to increase performance. We also
provide a freely available toolbox for the interested readers that would like to enter
this growing field of investigation. 

The reminder of this paper is as follows: in Section~ \ref{sec:reg}, we present a general  framework for regularized remote sensing image processing and discuss different forms of convex and non-convex regularization. \RR{}{We will also present the optimization algorithm proposed}. Then, in Section~\ref{sec:appl-remote-sens} we apply the proposed non-convex regularizers to the problem of multi- and hyper-spectral image classification \RR{}{and therefore present the specific data term for classification and study it in synthetic and real examples}. In Section~\ref{sec:sparse-line-unmix} we apply our proposed framework to the problem of linear unmixing, \RR{}{present the specific data term for unmixing} and study the behaviour of the different regularizers in simulated examples involving true spectra from the USGS library. Section~\ref{sec:conclusions} concludes the paper.

\section{Optimization and non convex regularization}
\label{sec:reg}

{
In this Section, we give an intuitive explanation of regularized models. We first
introduce the general problem of regularization and then explore convex and 
non-convex regularization schemes, with a focus on sparsity-inducing regularizers. 
Finally, we \RR{explicit}{present} the optimization algorithms to solve non-convex regularization,
with accent put on  proximal splitting methods such as GIST~\cite{gonggeneral}.}

\subsection{Optimization problem}

{Regularized models} address the following optimization problem:
\begin{equation}
  \label{eq:optprob}
\min_{\w\in\R^d} \quad L(\w) +\lambda R(\w)
\end{equation}
where $L(\cdot)$ is a smooth function (Lipschitz gradient), $\lambda>0$
is a regularization
parameter and $R(\cdot)$ is a regularization function. This kind of
problem is extremely common in data mining, denoising and parameter
estimation. 

$L(\cdot)$ is often an empirical loss that measure{s} the discrepancy
between a model $\w$ and a dataset containing real life
observations. 

The regularization term $R(\cdot)$ is added to the optimization
problem in order to promote a simple model, {which has
    been shown to lead to a
  better estimation} \cite{bousquet2002stability}. All the
regularization terms discussed in this work are of the form :
\begin{equation}
  \label{eq:1}
  R(\w)=\sum_k g(|w_k|)
\end{equation}
where $g$ is a monotonically increasing function. This means that the
complexity of the model $\w$ can be expressed as a sum of the
complexity of each {feature $k$ in the model.}

\begin{figure}[t]
  \centering
  \includegraphics[width=.95\columnwidth]{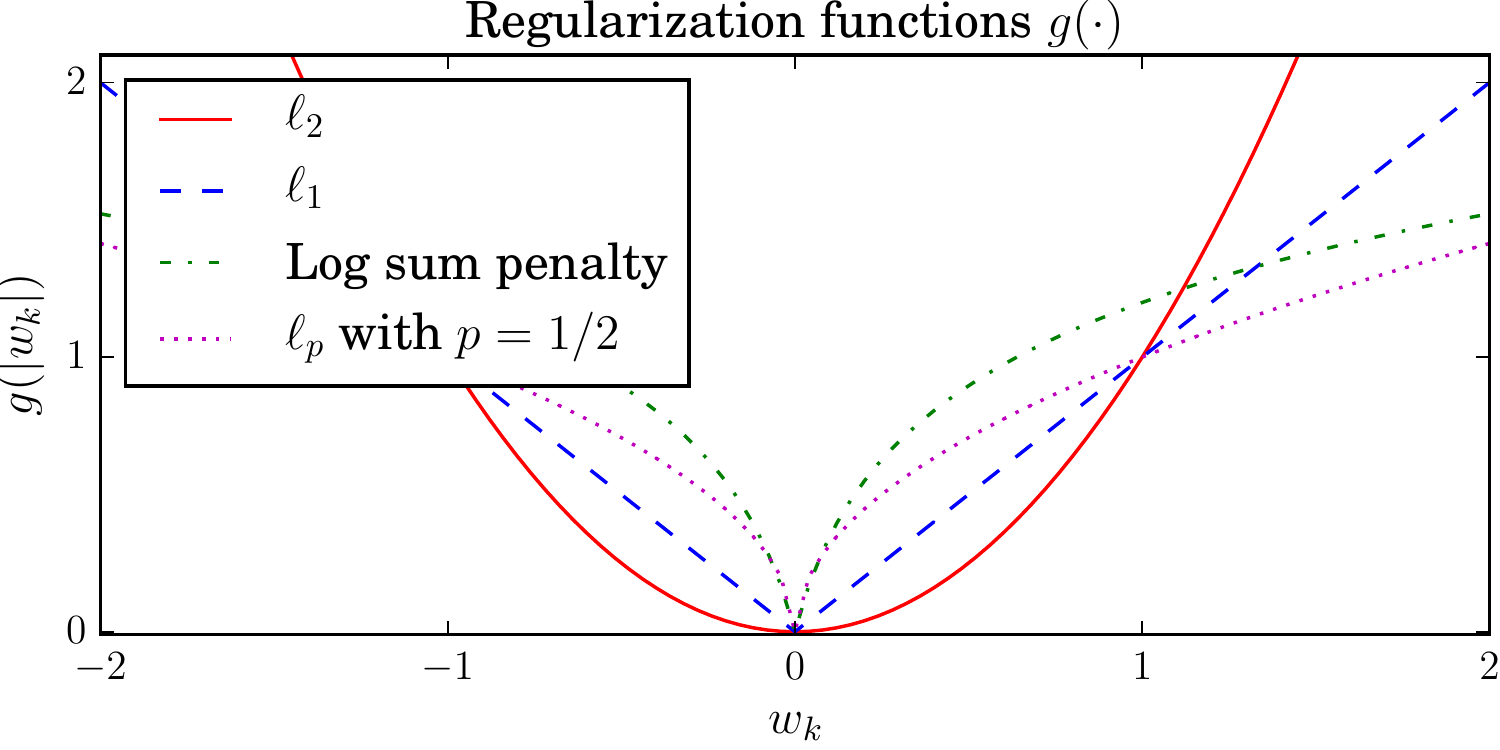}
  \caption{Illustration of the regularization terms $g(\cdot)$. Note
    that both $\ell_2$ and $\ell_1$ regularizations are convex and
    that log sum penalty and $\ell_p$ with $p=1/2$ are concave on
    their positive orthant.}
  \label{fig:visuregterm}
\end{figure}

The specific form of the regularizer will change the
  assumptions made on  the model. In the
  following, we discuss several classes of
  regularizers of increasing complexity:
  differentiable, non-differentiable (\emph{i.e.} sparsity inducing) and
finally both non-differentiable and non-convex. A summary of
all the regularization terms investigated in this work
is {given} in Table~\ref{tab:regterm}, along with an illustration of
the  regularization as a
  function of the value of the
  coefficient $w_k$
(Fig.~\ref{fig:visuregterm}). 

\begin{table}[t]
  \centering
  \caption{Definition of the regularization terms considered \label{tab:regterm}}
  \begin{tabular}{|l|c|}
    \hline
    Regularization term & $g(|w_k|)$\\\hline
    Ridge, $\ell_2$ norm & $|w_k|^2$\\
    Lasso, $\ell_1$ norm & $|w_k|$ \\
    Log sum penalty (LSP) & $\log(|w_k|/\theta+1)$ \\
    $\ell_p$ with $0<p<1$ & $|w_k|^p$ \\\hline
  \end{tabular}

\end{table}

\subsection{Non-sparse regularization}

One of the most common regularizers is 
the square $\ell_2$ norm of model $\w$,  \emph{i.e.}, $R(\w)=\|\w\|^2$
($g(\cdot)=(\cdot)^2$). This regularization will penalize large values
in the vector $\w$ but is isotropic, \emph{i.e.} it will not promote a
given direction for the vector $\w$. This regularization  term is also
known as $\ell_2$, quadratic or ridge regularization and is commonly
used in linear regression and classification. For instance, logistic
regression is often regularized with a quadratic term. Also note that
Support Vector Machine are regularized using
the $\ell_2$ norm in the Reproducing Kernel Hilbert Space of the form
$R(\w)=\w^\top\K\w$ \cite{scholkopf1998advances}. 
\subsection{Sparsity promoting regularization}

In some cases, not all the features or observations are of interest
for the model. In order to get a better estimation, one wants the
vector $\w$ to be sparse, \emph{i.e.} to have several components exactly
$0$.  For linear
  prediction, sparsity in the model $\w$ implies
that not all features are used for the prediction\footnote{\RR{}{Note that zero coefficients might happen also in the $\ell_2$ solution, but the regularizer itself does not promote their appearence.}}. This means that the
feature{s showing}
a non-zero value {in $w_k$} are then ``selected''. Similarly, when
estimating a
mixture one can suppose that only few materials are present, which
again implies sparsity {of the abundance coefficients}.

In order to promote sparsity in $\w$ one need\RR{}{s} to use a regularization
term that increases when the number of active component grows. The
obvious choice is to use the $\ell_0$ pseudo-norm that returns
directly the number of non-zero coefficients in $\w$. Nevertheless,
th\RR{is}{e $\ell_0$} term is non-convex and non-differentiable{, and cannot}  be
optimized exactly {unless} all the possible
subsets are tested.  Despite recent works aiming
 at solving directly this problem via discrete
  optimization \cite{bourguignon2016exact}, this approach is still 
computationally impossible even for medium-sized problems. \RR{O}{Greedy optimization
methods have been proposed to solve this kind of optimization problem
and have lead to efficient algorithms such as Orthogonal Matching
Pursuit (OMP) \cite{pati1993orthogonal} or Orthogonal Least Square
(OLS) \cite{chen1991orthogonal}. However, o}ne of the most common approaches to promote sparsity {without recurring
  to the $\ell_0$ regularizer} is to
use the $\ell_1$ norm \RR{}{instead}. This approach{, also} known as the Lasso
in linear
regression, has been  widely used in compressed sensing
in order to estimate with precision a few component in a large sparse
vector. 

\begin{figure}[t]
  \centering
  \includegraphics[width=1\linewidth]{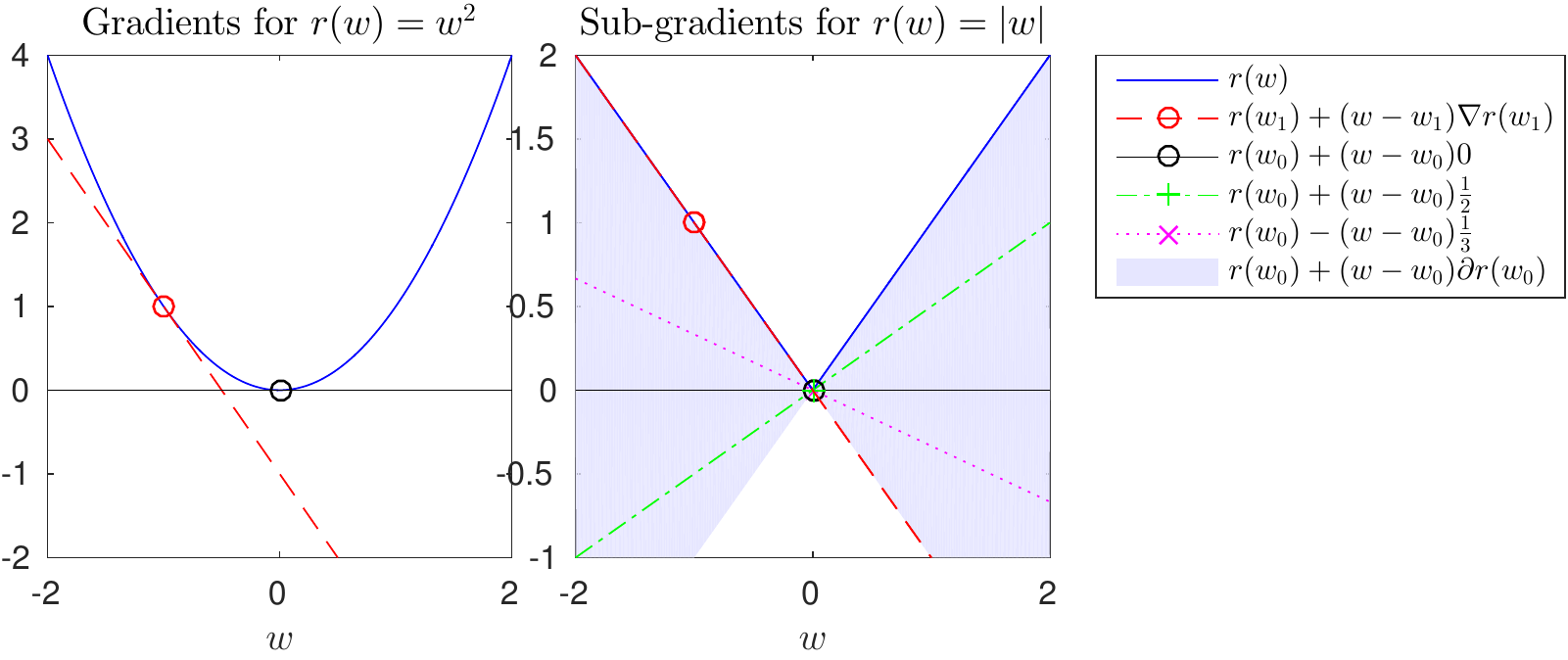}
  \caption{Illustration of gradients and subgradients on a
    differentiable {$\ell_2$} (left) and non-differentiable {$\ell_1$} (right) function. \label{fig:visusousgradient}}
\end{figure}

Now we discuss the intuition why using a regularization term such as
$\ell_1$ promotes sparsity. 
The reason {behind the sparsity of the $\ell_1$ norm} lies in the
non-differentiability at $0$ {shown} in Fig.~\ref{fig:visuregterm}
{(dashed blue line)}. 
For the sake of readability, we will suppose
here that $R(\cdot)$ is convex, but the intuition is the same and the
results can be generalized to the non-convex functions
{presented in the next section}. For a more illustrative
  example we use a 1D comparison between the $\ell_2$ and $\ell_1$ regularizers  (Fig.~\ref{fig:visusousgradient}).

\begin{itemize}
\item[-] When both the data and regularization term are
differentiable, a stationary point $\w^\star$ has the following
property: 
\begin{equation}\nabla L(\w^\star)+\lambda\nabla
R(\w^\star)=\mathbf{0}.\label{eq:statPoint}\end{equation} In other words, the gradients of both functions
have to cancel themselves exactly.
This is true for the $\ell_2$ regularizer everywhere, but also
  for the $\ell_1$, with the exception of $w_k = 0$. If we consider
  the $\ell_2$ regularizer as an example {(left plot in Fig.~\ref{fig:visusousgradient})}, we see that each point has a
  specific gradient, corresponding \RR{by}{to} the tangent to each point
  (e.g. the red dashed line). The stationary point is reached in this case for $w_k = 0$, as given by the black line in the left plot of Fig.~\ref{fig:visusousgradient}.

\item[-]  When the second {term in Eq~\eqref{eq:statPoint}} is not
  differentiable {(as in the $\ell_1$ case at $ 0$ presented in the
    right plot of Fig.~\ref{fig:visusousgradient})}, the gradient is
  not unique anymore and one has to use the sub-gradients and
  sub-differentials. For a convex function $R(\cdot)$ a sub-gradient
  at $\w^t$  is a vector $\x$ such that $R(\w)\geq\x^\top
  (\w-\w^t)+R(\w^t) $, \emph{i.e.} it is the slope of a linear
  function that remains below the function. In 1D, a sub-gradient
  defines a line touching the function at the non-differentiable point
  {(in the case of Fig.~\ref{fig:visusousgradient}, at 0)}, but
  stays below the function {everywhere else}, e.g. the  black
  and  green dotted-dash lines in   Fig.~\ref{fig:visusousgradient}
  (right). The sub-differential $\partial R(\w^t)$ is the set of all the
  sub-gradients that respect the minoration relation above. The
  sub-differential is   illustrated  in
  Fig.~\ref{fig:visusousgradient}, by the the area in light blue,
  which    contains all possible solutions.

Now the optimality constraints cannot rely
on equality since the sub-gradient is not unique, which leads to the
following optimality condition
\begin{equation}\mathbf{0}\in \nabla L(\w^\star)+\lambda\partial R(\w^\star)\end{equation}
This is very interesting in our case because {this condition} is much easier to satisfy {than Eq.~\eqref{eq:statPoint}}. Indeed, we just need to have a {single}
sub-gradient in  the whole sub-differential $\partial R(\cdot)$ that
can cancel the gradient $\nabla L(\cdot)$. {In other words, only one
  of the possible straight lines in the blue area is needed to cancel the
  gradient, thus making the chances for a null coefficient much
  higher.} For instance, when using the
$\ell_1$ regularization, the sub-differential of variable $w_i$ in $0$
is the set $[-\lambda,\lambda]$. When $\lambda$ becomes large enough it is
larger than all the components of the gradient $\nabla L(\cdot)$ and
the only solution verifying the conditions is the null vector
$\mathbf{0}$.

\end{itemize}

\begin{figure}[t]
  \centering
  \includegraphics[width=\linewidth]{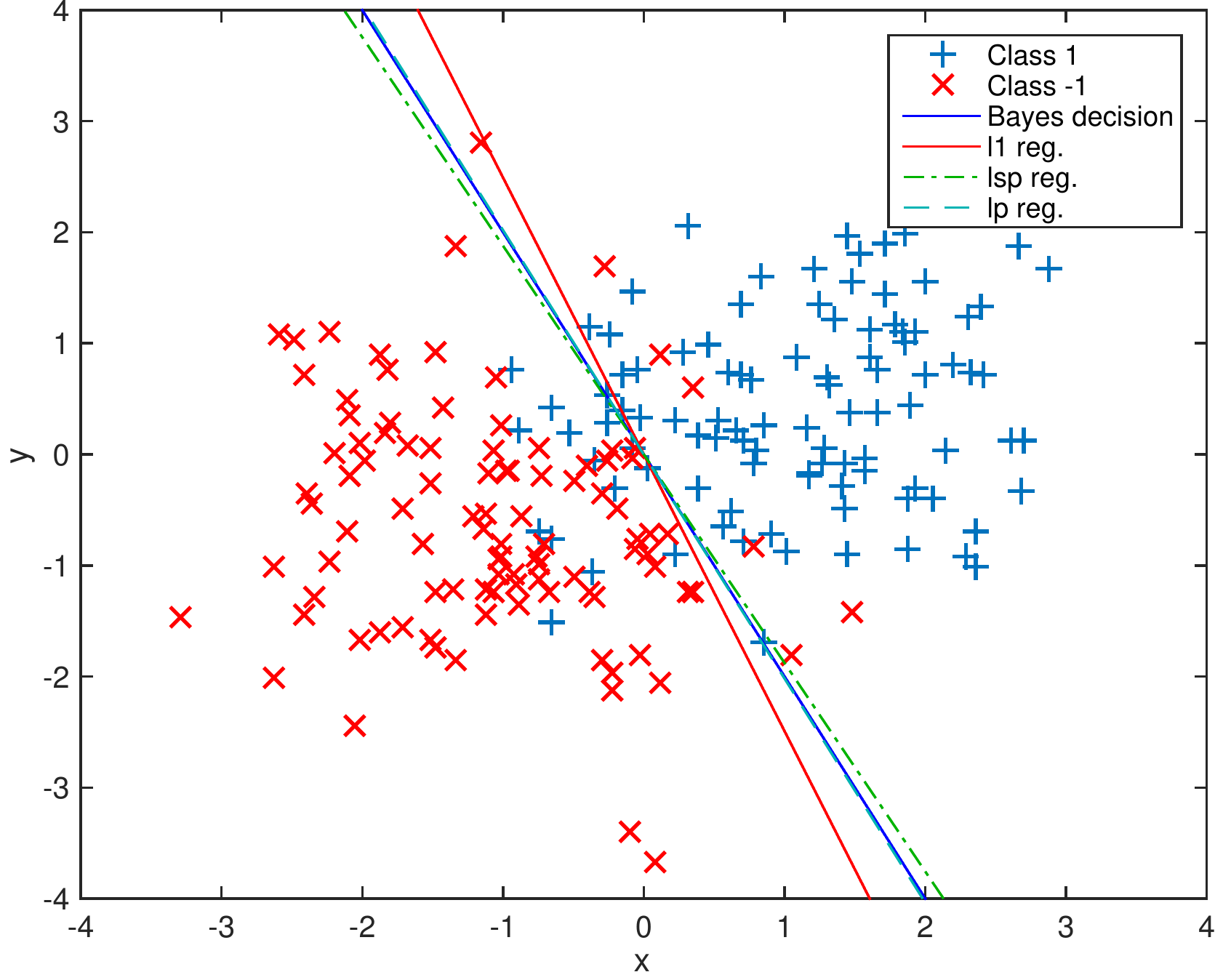}
  \caption{Example for a 2-class toy example with 2 discriminant
    features and 18 noisy features. The regularization parameter of
    each method has been chosen as the minimal value that leads to the
    correct sparsity with only 2 features selected. \label{fig:toybias}}
\end{figure}

The $\ell_1$ regularization has been largely studied. Because it is
convex meaning it avoids the problem of local minima, and many
efficient optimization procedures exists to solve it (e.g. LARS
\cite{efron2004least}, Forward
Backward Splitting \cite{combettes2011proximal}). But the sparsity of
the solution using $\ell_1$ regularization often comes with a cost in
term of generalization. While theoretical studies show that under
some constraint the Lasso can recover the true relevant
variables and their sign, the solution obtained will be biased
toward $\mathbf{0}$
\cite{zou2006adaptive}. Figure~\ref{fig:toybias} illustrates
  the bias in a two-class toy dataset: the $\ell_1$ decision function
  (red line) is biased with respect to the Bayes decision function
  (blue line). In this case,  the bias corresponds to a rotation of the separating hyperplane. In practice, one can
deal with this bias by estimating again the model on selected subset
of variables {using an isotropic norm
  (e.g. $\ell_2$)}~\cite{tuia2014automatic}, but this requires to
solve again
an optimization problem. {The approach we
  propose in this paper} is to use a non-convex
regularization term that will still promote sparsity, while minimizing
the aforementioned bias{. To this end, we
  present non-convex regularization in the next section}.

\subsection{Non-convex regularization}

In order to promote more sparsity while reducing the bias, several
works have looked at non-convex, yet continuous regularization. {Such regularizers} have been proposed for instance in statistical
estimation  \cite{fan2001variable},
 compressed sensing
\cite{candes2008enhancing} or in machine learning
\cite{zhang2010analysis}. 
Popular examples are the Smoothly Clipped Absolute Deviation
(SCAD) \cite{fan2001variable}, the Minimax Concave Penalty (MCP)
\cite{zhang2010nearly} and the Log Sum Penalty (LSP)
\cite{candes2008enhancing} {considered below} (see \cite{gonggeneral} for more
examples). In the following we will investigate two of them \RR{more
in detail}{in more detail}: $\ell_p$ pseudo-norm with $p=\frac{1}{2}$
and LSP, both also displayed in Figure \ref{fig:visuregterm}.

 All the non-convex regularization  above share some
 particular characteristics that make them of interest in our
 case. First {(and as the $\ell_0$ pseudo-norm and  $\ell_1$ norm)}
 they all have a non-differentiability in  $\mathbf{0}$,
 which {-- as we have seen in the previous section --} promotes
 sparsity. Second they are all concave in their positive
 orthant, which limits the bias because their gradient will decrease
 for large values of $w_i$ limiting the shrinkage (as compared to the
 $\ell_1$ norm, whose gradient for
 $w_i\neq 0$  is constant). {Intuitively, this means that with
   a non-convex regularization it will become more difficult for large
   coefficients to be shrinked toward 0, because their gradient is small. On the
   contrary, the $\ell_1$ norm will treat all coefficients equally and
   apply the same attraction to the stationary point to all
   of them. {The decision functions for the LSP and
   $\ell_p$ norms are shown in Fig.~\ref{fig:toybias} and are much
   closer to the actual (true) Bayes decision function.}

\subsection{Optimization algorithms}
\label{sec:optim-algor}

Thanks to the differentiability of the $L(\cdot)$ term, the
optimization problem can be solved using proximal splitting methods
\cite{combettes2011proximal}. The convergence of those
algorithms to a global minimum are well studied in the convex case. For
non-convex regularization, recent works have proved that proximal
methods can be used with non-convex regularizers when a simple closed
form solution of the proximity operator for the regularization can be
computed~\cite{gonggeneral}. Recent works have studied the convergence
of proximal methods with non-convex regularization and proved
convergence to a local stationnary
point   for
a large family of loss functions \cite{attouch2010proximal}.

In this work, we used the General Iterative Shrinkage
and Thresholding (GIST) algorithm proposed in~\cite{gonggeneral}. This
approach is a first order method that consists in iteratively
linearizing $L(\cdot)$ in order to solve very simple proximal
operators at each iteration. At each iteration $t+1$  one
computes the model update $w^{t+1}$ by solving
\begin{equation}
  \label{eq:fbs}
  \min_{\w} \quad \nabla L(\w^t)^\top(\w-\w^t)+\lambda R(\w)+\frac{\mu}{2}\|\w-\w^t|\RR{}{|}^2_2.
\end{equation}
When $\mu$ is a Lipschitz constant of $L(\cdot)$,
the cost function above is a majorization of $L(\cdot)+\lambda
R(\cdot)$ which ensures a decrease of the objective function at each
iteration.  Problem \eqref{eq:fbs} can be reformulated as a proximity
operator
\begin{equation}
  \label{eq:prox}
 \text{prox}_{\lambda R}(\v)=\text{arg}\min_\w \quad \lambda R(\w)+\frac{\mu}{2}\|\w-\v^t\|_2^2,
\end{equation}
where $\v^t=\w^t-\frac{1}{\mu}\nabla L(\w^t)$ can be seen as a gradient
step \emph{w.r.t.} $L(\cdot)$  followed by a proximal operator at each
iteration. Note that the efficiency of a proximal algorithm depends
on {the existence of} a simple closed form solution for solving the
proximity operator
in Eq.~\eqref{eq:prox}. \RR{}{ Luckily, there exists numerous operators in the convex
case (detailed list in \cite{combettes2011proximal}) and some non-convex
proximal operator can be computed on the regularization used in our
work (see \cite[Appendix 1]{gonggeneral} for LSP and \cite[Equ. 11]{xu2012regularization}
for $\ell_p$ with $p=1/2$)}. Note that  efficient methods, which
estimate the Hessian matrix
\cite{chouzenoux2013block,rakoto2015dcprox} exist, as well as  a wide range of
methods based on DC programming, which have shown to work very well in
practice \cite{rakoto2015dcprox,tnnls2014} and \RR{}{can handle the general
case $p\in(0,1]$ for the $\ell_p$ pseudo-norm (see~\cite{gasso2009recovering} for an implementation)}.

Finally, when one wants to perform variable selection using the
$\ell_0$ pseudo-norm as
regularization, the exact solution of the combinatorial
  problem is not always necessary.  \RR{G}{As mentioned above,} greedy optimization
methods have been proposed to solve this kind of optimization problem
and have lead to efficient algorithms such as Orthogonal Matching
Pursuit (OMP) \cite{pati1993orthogonal} or Orthogonal Least Square
(OLS) \cite{chen1991orthogonal}. In this paper, we won't
  consider these methods in detail, but they have been shown to perform
  well on least square minimization problems.

\section{Classification with feature selection}
\label{sec:appl-remote-sens}

In this section, we tackle the problem of sparse
classification. Through a toy example and a series of real data
experiments, we will
study the interest of non convex regularization.

\subsection{Model}
\label{sec:model}
The model we will consider in the experiments is a simple linear classifier of the form $f(\x)=\w^\top\x+b$
where $\w\in\mathbb{R}^d$ is the normal vector to the separating
hyperplane and $b$ is a bias term. In the binary case ($y_i \in [-1;1]$), the estimation is performed by
solving the following regularized optimization problem:
\begin{equation}
  \label{eq:optprob}
  \min_{\w,b}\quad \frac{1}{n}\sum_{i=1}^n \mathcal{L}(y_i,f(\x_i)) + R(\w),\end{equation}
where $R(\w)$ is one of the regularizers in Tab\RR{.}{le}~\ref{tab:regterm} and $\mathcal{L}(y_i,f(\x_i))$ is a classification loss that measures the
discrepancy between the prediction $f(\x_i)$ and the true label $y_i$. Hereafter, we will use the squared hinge loss:
$$\mathcal{L}(y_i,f(\x_i)) = \max(0,1-y_if(\x_i))^2.$$

When dealing with multi-class problems, we use a One-Against-All procedure, i.e. we learn one linear
function $f_k(\cdot)$ per class $k$ and then predict the final class for a given
observed pixel $\x$ as the solution of $ \arg\min_k f_k(\x)$. In
practice, this leads to an optimization problem similar to
Eq.~\eqref{eq:optprob}, where we need to estimate a matrix $\W$, containing the coefficients per each class. The number of coefficients to be estimated is
therefore the size $d$ of the input space multiplied by the number of classes $C$.

\subsection{Toy example}
\label{sec:toy-example}
First we consider in detail the toy example in Fig.~\ref{fig:toybias}:
the data considered are 20-dimensional, where the first two dimensions
are discriminative (they correspond to those plotted in
Fig.~\ref{fig:toybias}), while the other are not (they are generated
as Gaussian noise). The correct solution is therefore to assign non
zero coefficients to the two discriminative features and $w_k = 0$ for
all the others.

Figure~\ref{fig:toybias} show the classifiers estimated for
  the smallest value of the regularization term $\lambda$, which leads
  to the correct sparsity level (2 features selected). This ensures that we
  have selected the proper components, while minimizing the bias for all
  methods.
 This also illustrates that the $\ell_1$ classifier has
a stronger bias (i.e. provides a decision function further away from
the optimal Bayes classifier) than the classifiers regularized by
non-convex functions. 

Let's now focus on the effect of the
regularization term and of its strength, defined by the regularization
parameter $\lambda$ in
Eq.~\eqref{eq:optprob}. Figure~\ref{fig:toyregpath} illustrates a
regularization path, i.e., all the solutions obtained by increasing the
regularization parameter $\lambda$\footnote{A ``regularization path'' for the Lasso is generally computed using homotopy algorithms \cite{fht}.
However, experiments show that the computational complexity of the complete Lasso path remains \RR{}{high} for
high-dimensional data. Therefore, in our experiments we used an approximate path (i.e., a discrete sampling of $\lambda$ values along the path).}. Each line corresponds to one input
variable and those with the largest coefficients (and in color) are the
discriminative ones. Considering the $\ell_2$ regularization (top left
panel in Fig.~\ref{fig:toyregpath}), no sparsity is achieved and, even
if the two correct features have the largest coefficients, the
solution is not compact. The $\ell_1$ solution (top right panel) shows
a correct sparse solution for $\lambda = 10^{-1}$ (vertical black
line, where all the coefficients but two are $0$), but the smallest
coefficient is biased (it is smaller than expected by the Bayes
classifier, represented by the horizontal dashed lines). The two
non-convex regularizers (bottom line of Fig.~\ref{fig:toyregpath})
show the correct features selected, but a smaller bias: the
coefficient retrieved are closer to the optimal ones of the Bayes
classifier. Moreover, the non zero coefficients stay close to the
correct values for a wider set of regularization parameters and then
drop directly to zero: this means that the non-convex model either has
not enough features to train or has little feature with the right
coefficients, contrarily to the $\ell_1$ that can retrieve sparse
solution with wrong coefficients, as it can be seen in the
part to the right of the vertical black line of the $\ell_1$
regularization path.

\begin{figure}
\includegraphics[width=\linewidth]{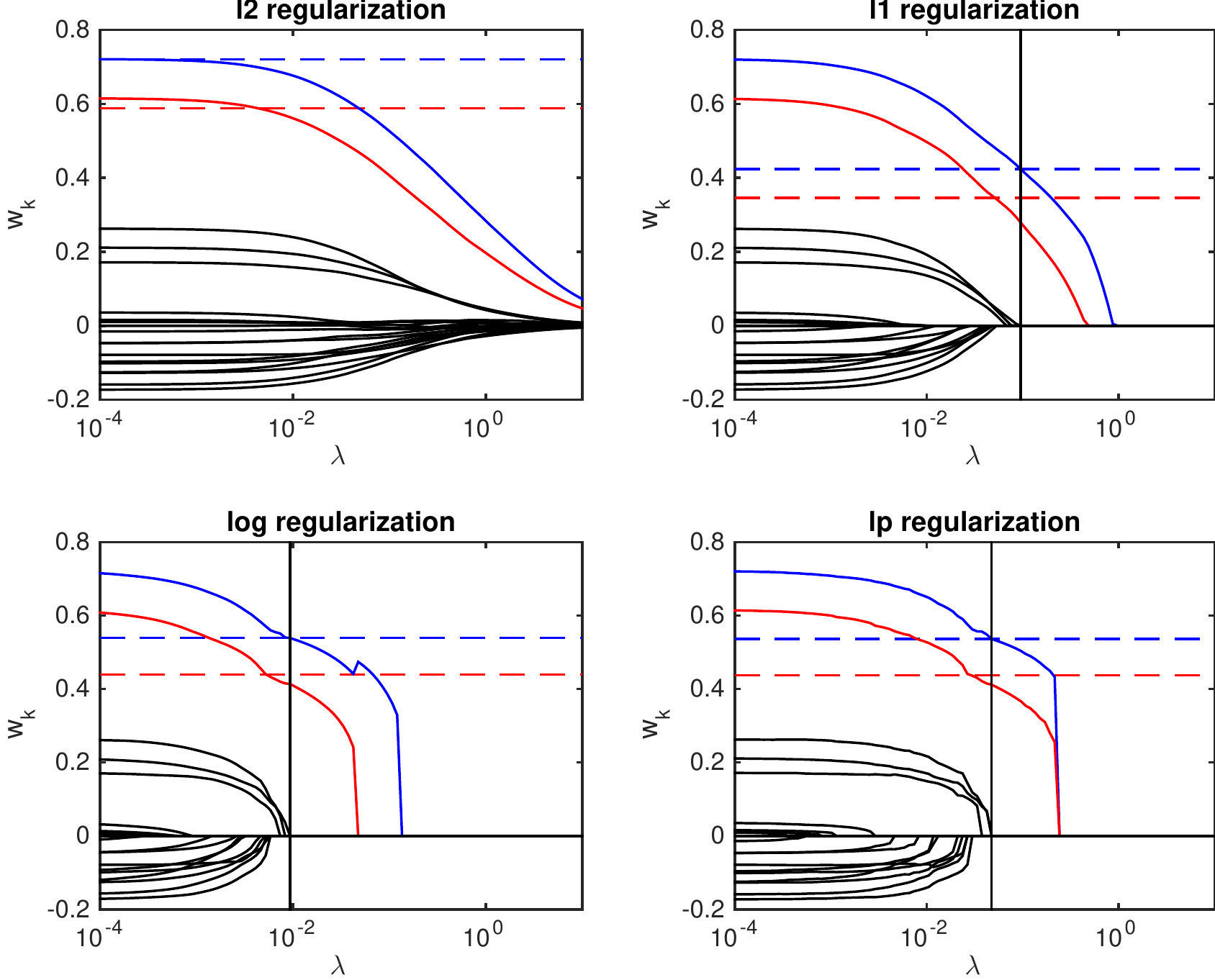}
\caption{Regularization paths for the toy example in Fig.~\ref{fig:toybias}. Each line corresponds to the coefficients $w_k$ attributed to each feature along the different values of $\lambda$. The best fit is met for each regularizer at the black vertical line, where all coefficients but two are $0$. The unbiased Bayes classifier coefficients (the correct coefficients) are represented by the horizontal dashed lines.\label{fig:toyregpath}}
\end{figure}

\subsection{Remote sensing images}
\label{sec:real-life-dataset}

\noindent{\textbf{Data.}} The real datasets considered are three very high resolution remote sensing images.

\begin{enumerate} 
\item \RR{}{\textsc{Thetford Mines}. The first dataset is acquired over the Thetford mines site in Qu\'ebec, Canada and contains two data sources: a VHR color image (three channels, red-green-blue) at 20 cm resolution and a long wave infrared (LWIR, 84 channels) hyperspectral image at approximatively 1 m resolution\footnote{The data were proposed as the Data Fusion Contest 2014~\cite{Lia14} and are available on the IADF TC website for download \href{http://www.grss-ieee.org/community/technical--committees/data--fusion/}{http://www.grss-ieee.org/community/technical-committees/data-fusion/}}. The LWIR images are downsampled by a factor 5, to match the resolution of the RGB data, leading to a $(4386 \times 3769 \times 87)$ datacube. The RGB composite, band 1 of the LWIR data and the train / test ground truths are provided in Fig.~\ref{fig:mines14}.}
\item \textsc{Houston}. The second image is a CASI image acquired over Houston with $144$ spectral bands at $2.5m$ resolution. A field survey is also available ($14`703$ labeled pixels, divided in 14 land use classes). A LiDAR DSM was also available and was used as an additional feature\footnote{The data were proposed as the Data Fusion Contest 2013~\cite{grss_dataset} and are available on the IADF TC website for download \href{http://www.grss-ieee.org/community/technical--committees/data--fusion/}{http://www.grss-ieee.org/community/technical-committees/data-fusion/}}. The CASI image was corrected with histogram matching for a large shadowed part on the right side (as in~\cite{Tui15}) and the DSM was detrended by a 3m trend on the left-right direction. Image, DSM and ground truth are illustrated in Fig.~\ref{fig:houston}.
\item \textsc{Zurich Summer}. The third dataset is a series of $20$ QuickBird images acquired over the city of Zurich, Switzerland, in August 2002\footnote{The dataset is freely available at \href{https://sites.google.com/site/michelevolpiresearch/data/zurich-dataset/}{https://sites.google.com/site/ michelevolpiresearch/data/zurich-dataset}}. The data have been pansharpened at 0.6 $m$ spatial resolution and a dense ground truth is provided for each image. Eight classes are depicted: buildings, roads, railway, water, swimming pools, trees, meadows and bare soil. More information on the data can be found in~ \cite{Volpi2015b}. To reduce computational complexity, we extracted a set of superpixels using the Felzenszwalb algorithm~\cite{Felzenszwalb2004}, which reduced the number of samples from $\sim10^6$ pixels per image to a few thousands. An example of the superpixels extracted on image tile \#3 is given in Fig.~\ref{fig:felz}.

\end{enumerate}

\begin{figure}
	\begin{tabular}{cc}
		(a) RGB & (b) LWIR band 1 \\
		\includegraphics[width=.4\columnwidth]{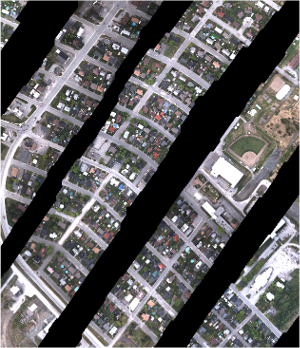} &
		\includegraphics[width=.4\columnwidth]{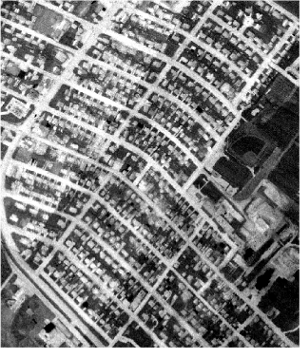} \\
		\includegraphics[width=.4\columnwidth]{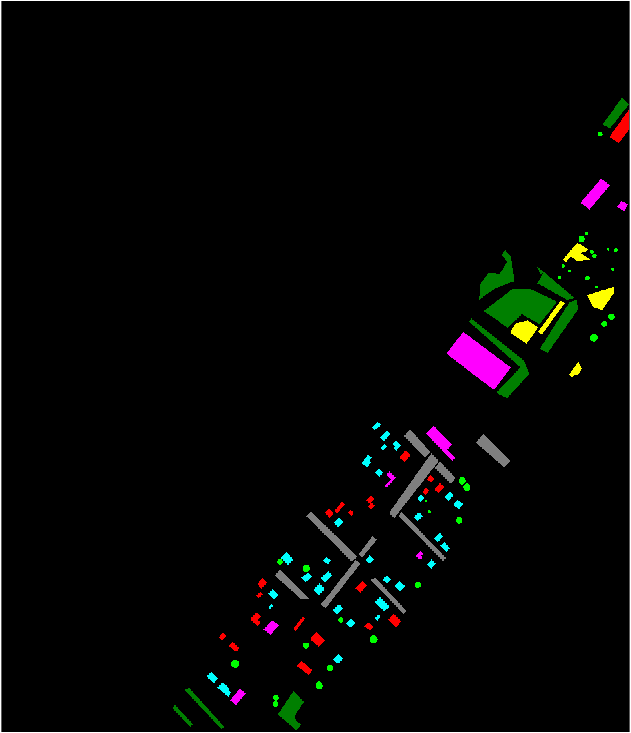} &
		\includegraphics[width=.4\columnwidth]{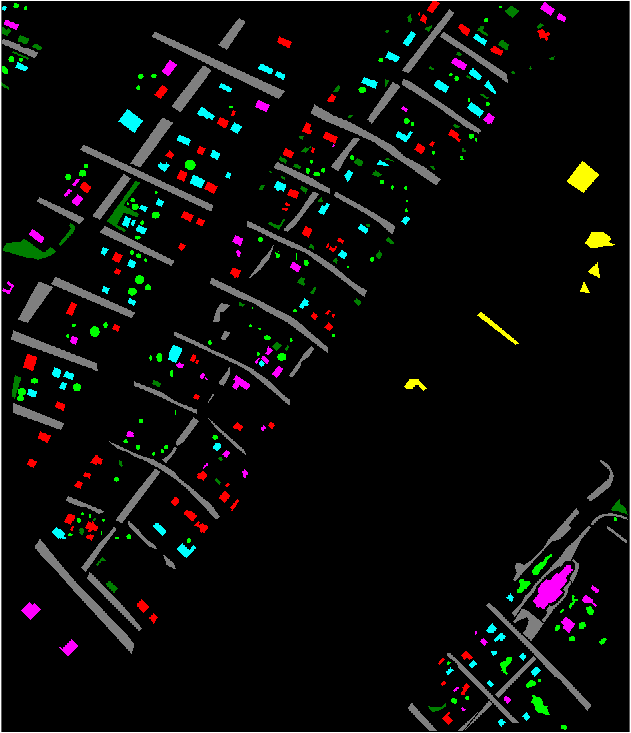} \\
		(c) GT training & (d) GT test \\
	\end{tabular}
\caption{\RR{}{The \textsc{Thetford mines 2014} dataset used in the classification experiments, along with its labels. \label{fig:mines14}}}	
\end{figure}

\begin{figure}
	\begin{tabular}{c}
		\includegraphics[width=.98\columnwidth]{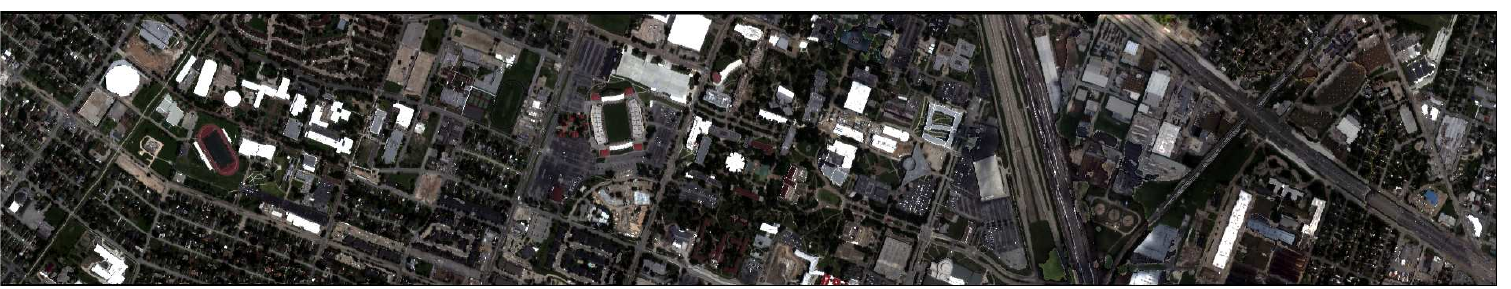} \\
		\includegraphics[width=.98\columnwidth]{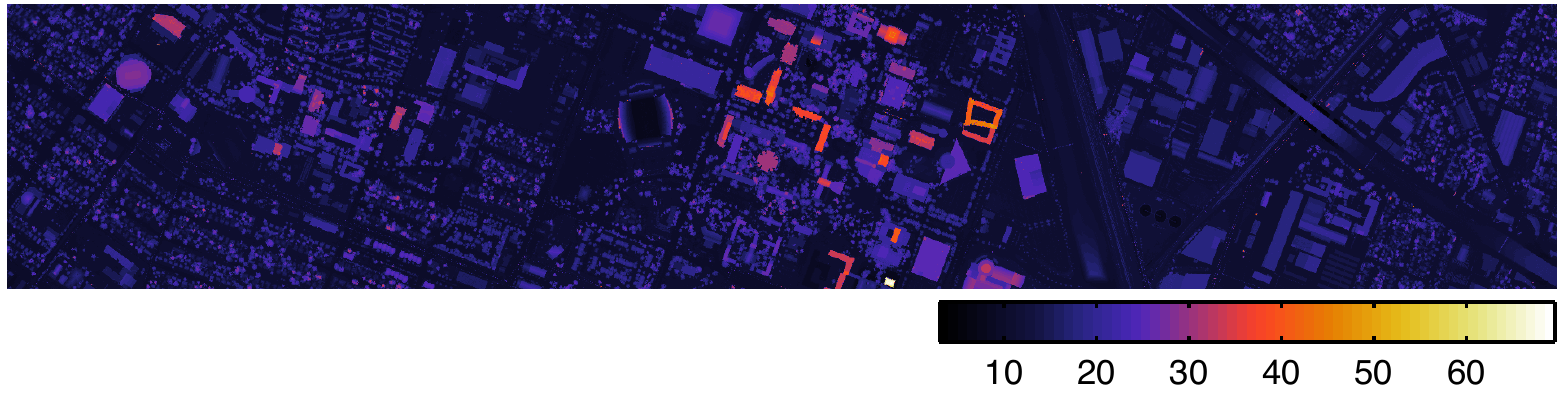} \\
		\includegraphics[width=.98\columnwidth]{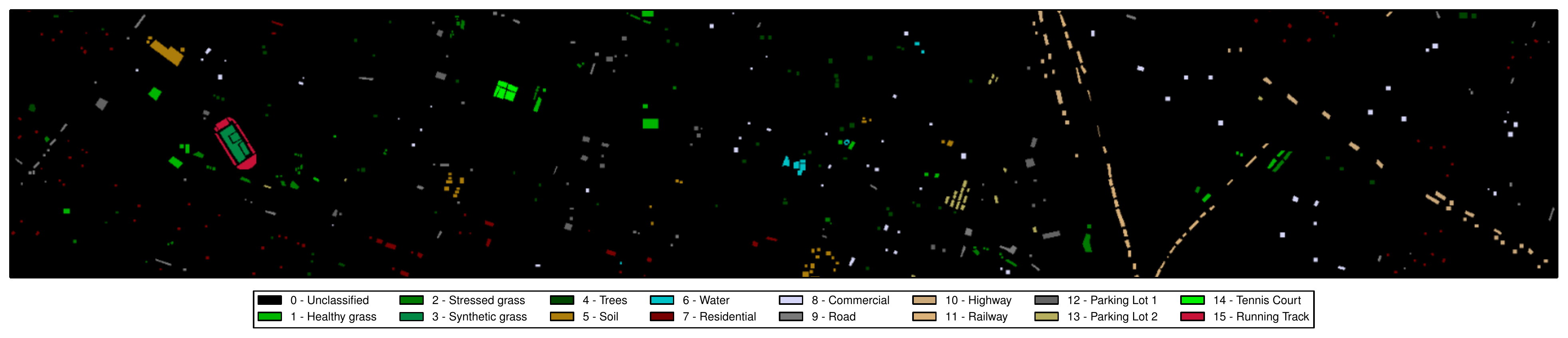}\\
	\end{tabular}
\caption{The \textsc{Houston} dataset used in the classification experiments: (top) true color representation of the hyperspectral image (144 bands); (middle): detrended LiDAR DSM; (bottom) labeled samples (all the available ones, in 15 classes). \label{fig:houston}}	
\end{figure}

\begin{figure}
\setlength{\tabcolsep}{1pt}
\begin{tabular}{cc}
\includegraphics[width = .23\textwidth]{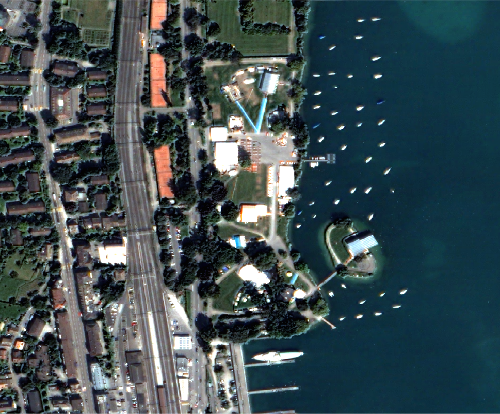}&
\includegraphics[width = .24\textwidth]{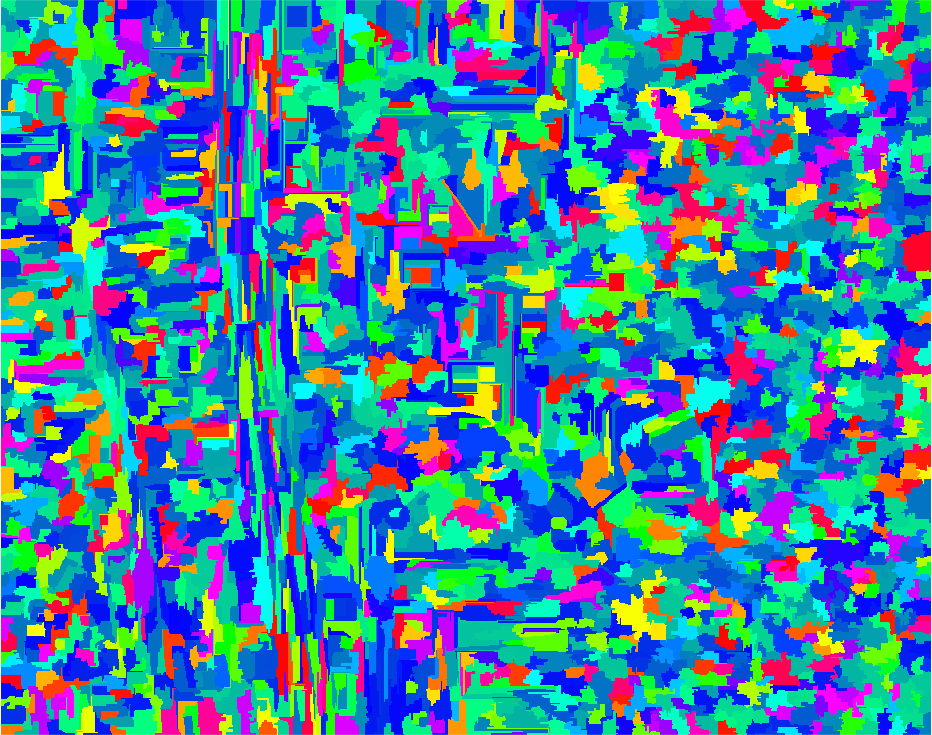}\\
\end{tabular}
\caption{Example on tile \#3 of the superpixels extracted by the Felzenszwalb algorithm~\cite{Felzenszwalb2004}.\label{fig:felz}}
\end{figure}

\noindent{\textbf{Setup.}} For all datasets, contextual features were added to the spectral bands, in order to improve the geometric quality of classification~\cite{Fauvel13}:  morphological and texture filters were added, following the list in~\cite{tuia2014automatic}. \RR{}{Each image was processed to extract the most effective filters for its processing: }

\begin{itemize}
\item \RR{}{For the \textsc{Thetford mines} dataset, the filters were extracted from the RGB image and from a normalized ratio between the red band and the average of the LWIR bands (following the strategy of the winners of the 2014 Data Fusion Contest~ \cite{Lia14}), which approaches a  vegetation index. Given the extremely high resolution of the dataset, the filters were computed with the size range \{7, ... 23\}, leading to 100 spatial features.}
\item \RR{}{For the \textsc{Houston} case, the filters were calculated on both the $3$ first principal components projections extracted from the hyperspectral image and the DSM. Given the smaller resolution of this dataset, the convolution sizes of the local filters are in the range $\{3, ..., 15\}$ pixels. This leads to 240 spatial features.}
\item For the \textsc{Zurich summer} dataset spatial filters were computed directly on the four spectral bands, plus the NDVI and the NDWI indices. \RR{}{Then, average, minimum, maximum and standard deviation values per superpixel were extracted as feature values. Since the spatial resolution is comparable to the one of the \textsc{Houston} dataset, the same sizes of convolution filters are used, leading to a total of 360 spatial features.}
\end{itemize}

\RR{}{The joint spatial-spectral input space is obtained by stacking the original images to the spatial filters above. It is therefore of dimension $188$ in the \textsc{Thetford mines} data, $384$ in the \textsc{Houston} data and $366$ in the \textsc{Zurich summer} case.
}
\begin{figure*}\centering
\setlength{\tabcolsep}{2pt}
\begin{tabular}{ccccc}
\includegraphics[height = 2.5cm]{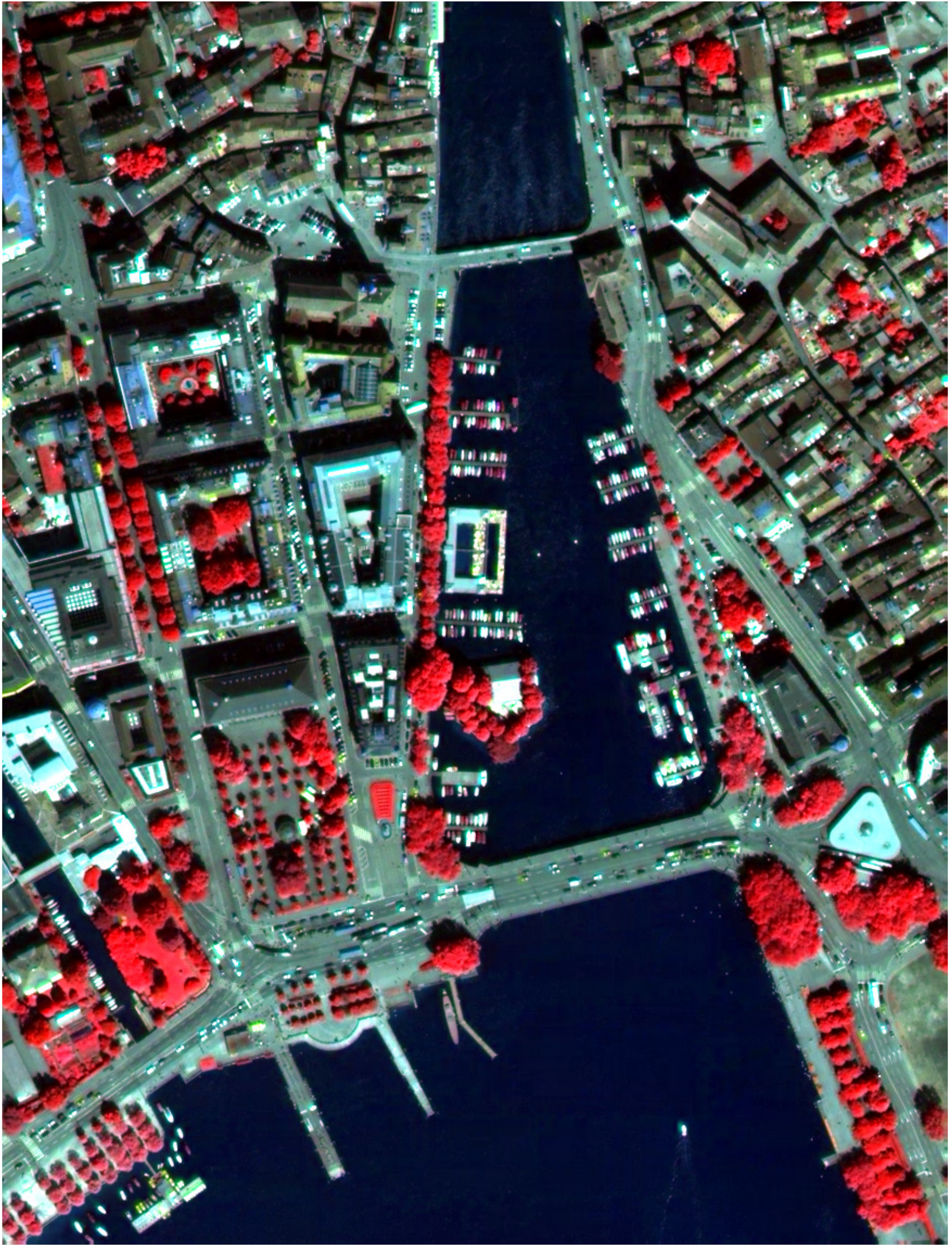}&
\includegraphics[height = 2.5cm]{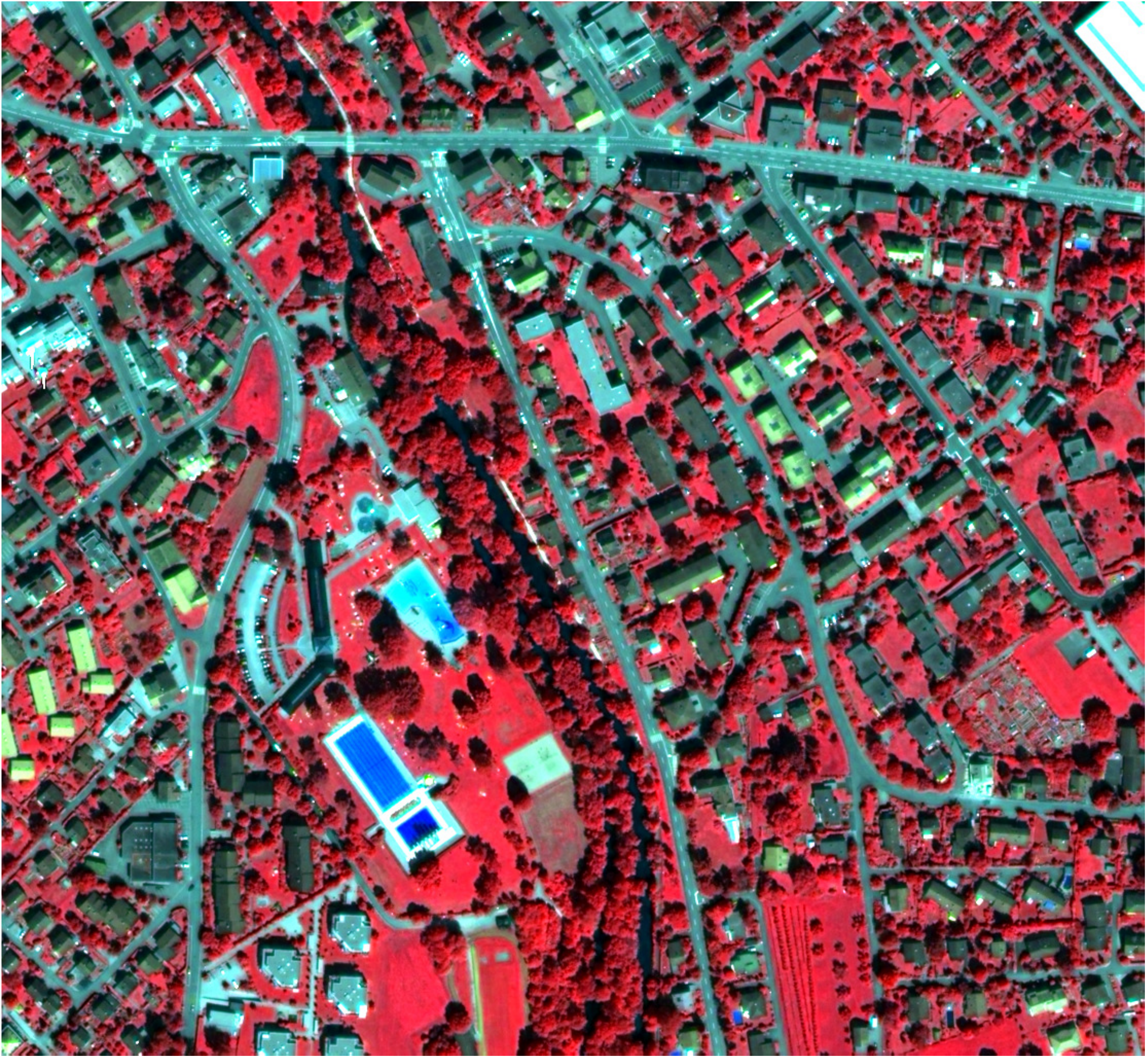}&
\includegraphics[height = 2.5cm]{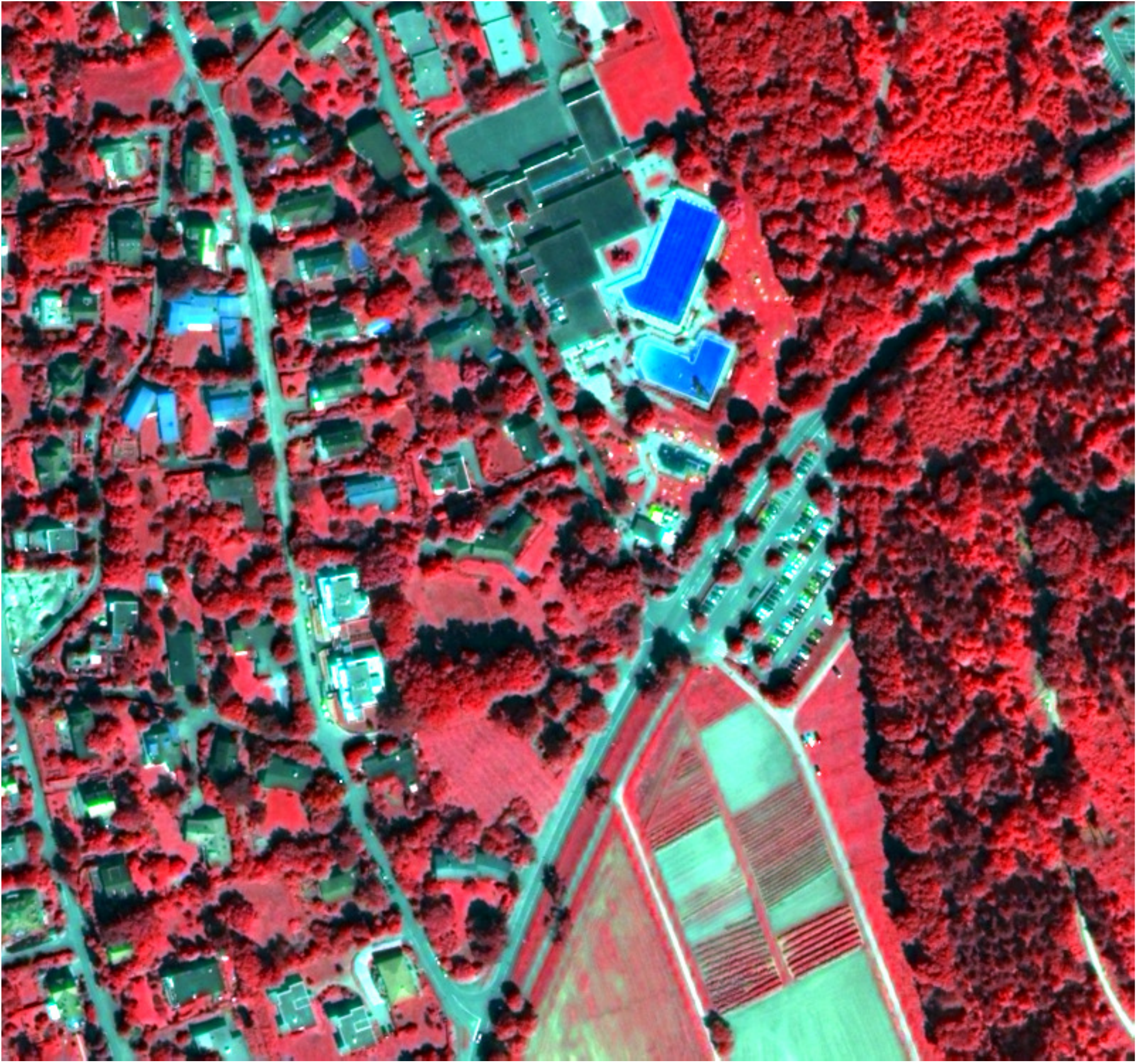}&
\includegraphics[height = 2.5cm]{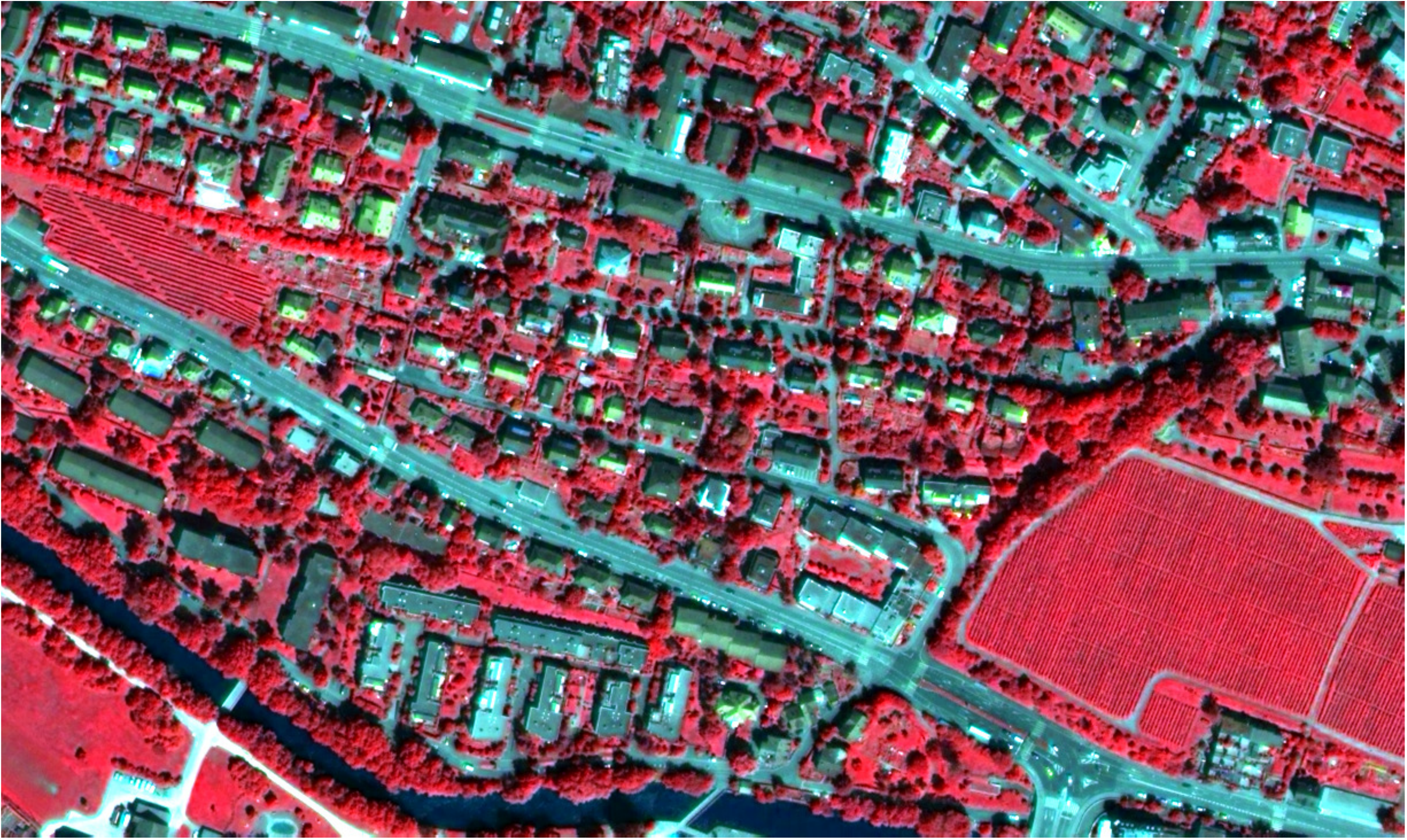}&
\includegraphics[height = 2.5cm]{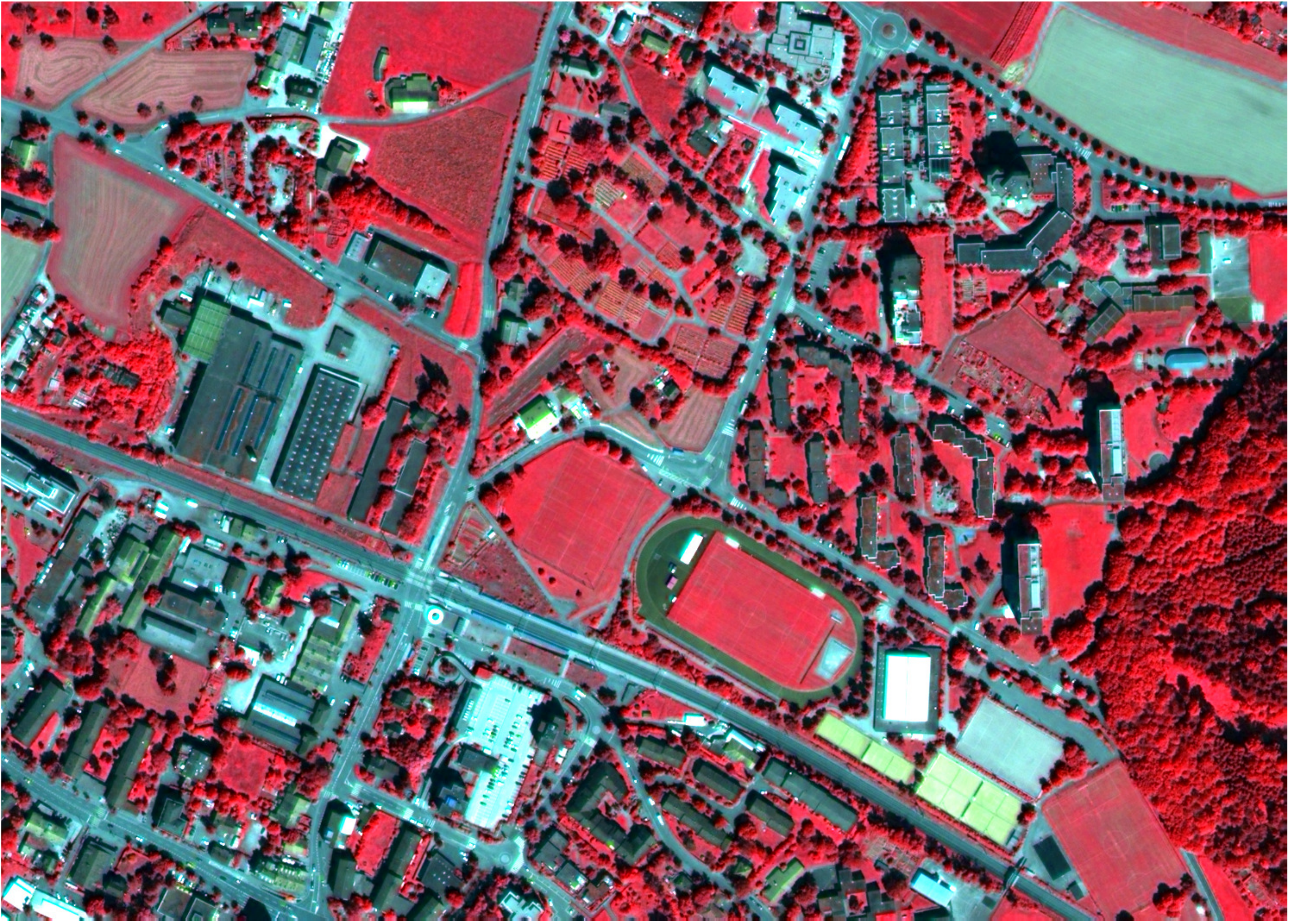}\\
\includegraphics[height = 2.5cm]{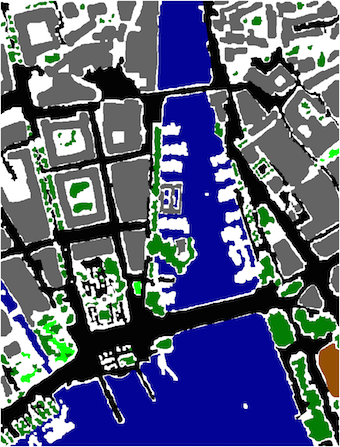}&
\includegraphics[height = 2.5cm]{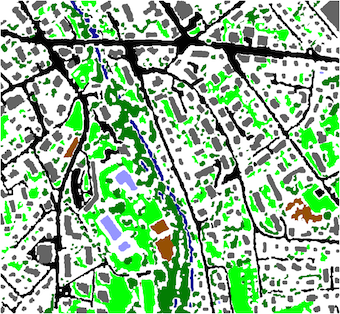}&
\includegraphics[height = 2.5cm]{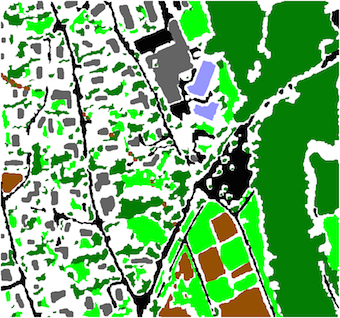}&
\includegraphics[height = 2.5cm]{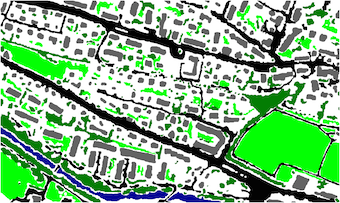}&
\includegraphics[height = 2.5cm]{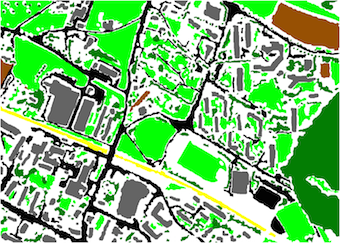}\\
\end{tabular}
\caption{The five test images of the Zurich Summer dataset (from left to right, tiles \#16 to \#20), along with their ground truth.\label{fig:zhTest}}
\end{figure*}

Regarding the classifier, we considered the linear classifier of Eq.~\eqref{eq:optprob} with a squared hinge loss:
\begin{itemize}
\item \RR{}{In the \textsc{Thetford mines} case, we use $5000$ labeled pixels per class. Given the spatial resolution of the image and the $568`242$ labeled points available in the training ground truth, this only represents approximatively 5\% of the labeled pixels in the training image. For test, we use the entire test ground truth, which is spatially disconnected to the training one (except for the class `soil', see Fig.~\ref{fig:mines14}) and carries $1.5$ million labeled pixels.}
\item In the \textsc{Houston} case, we also proceed with pixel classification. All the models are trained with 60 labeled pixels per class, randomly selected, and all the remaining labeled pixels are considered as the test set. We report performances on the  entire test set provided in the Data Fusion contest 2013\RR{}{, which is spatially disconnected from the training set (Fig.~\ref{fig:houston})}.

\item For the \textsc{Zurich Summer} data, we deal with superpixels and 20 separate images. We used images \#1-15 to train the classifier and then tested on the five remaining images (Fig.~\ref{fig:zhTest}). Given the complexity of the task (not all the images have all the classes and the urban fabrics depicted vary from scene to scene), we used 90\% of the available superpixels in the $15$ training images, which resulted in $30`649$ superpixels. All the labeled superpixels in the test images ($8`960$ superpixels) are used as test set.

\end{itemize}

\vspace{.3cm}

Regarding the regularizers, we compare the four regularizers of Tab.~\ref{tab:regterm} ($\ell_1$,
$\ell_2$, Log sum penalty and $\ell_p$ with $p = 1/2$) and study the
joint behavior of accuracy and sparsity along a regularization
path, i.e. for different values of $\lambda$: below 
$\lambda = \{ 1e^{-5},
\ldots, 1e^{-1} \}$, with 18 steps. For each step, the experiment was
repeated ten times with different train/test sets (each run with the same training samples for all regularizers) and the average Kappa and number of active coefficients is reported in Fig~\ref{fig:res}. Also note that we
report the total number of coefficients in the multiclass case,
$w_{j,k}$, which is equal to the number of features multiplied by the
number of classes, plus one additional feature per class (bias
term). In total, the model estimates \RR{}{$1`504$
coefficients in the case of the \textsc{Thetford mines} data}, while for the \textsc{Houston} and  \textsc{Zurich Summer} cases it deals with $5`775$ and $3`294$ coefficients, respectively.

\begin{figure*}
\begin{tabular}{ccc}
\setlength{\tabcolsep}{0pt}
\rotatebox{0}{\hspace{.2cm} \textsc{\RR{}{Thetford mines}}} & \rotatebox{0}{\hspace{.2cm} \textsc{Houston}} & \rotatebox{0}{\hspace{.2cm} \textsc{Zurich Summer}} \\
\includegraphics[width=.3\textwidth]{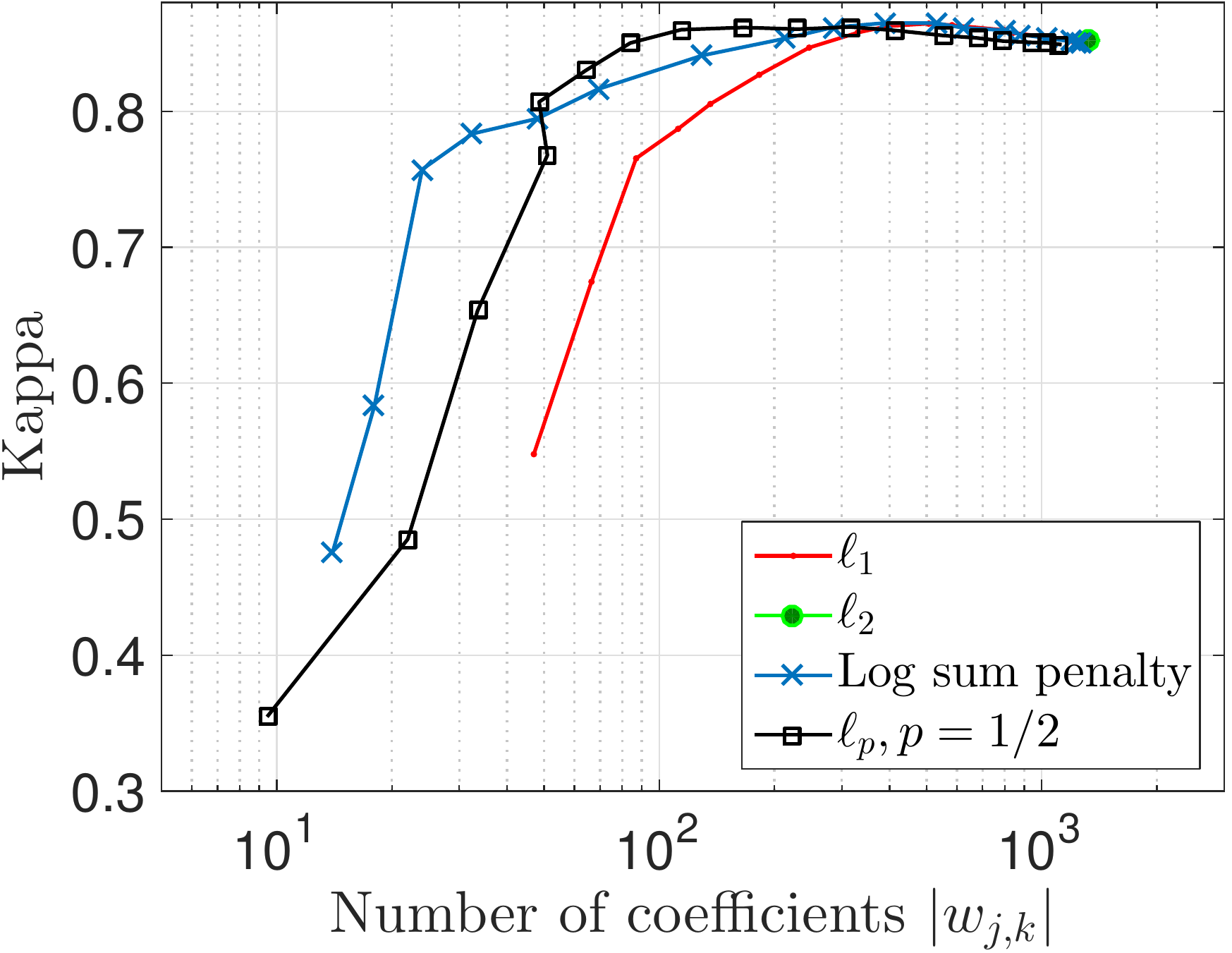} &
\includegraphics[width=.3\textwidth]{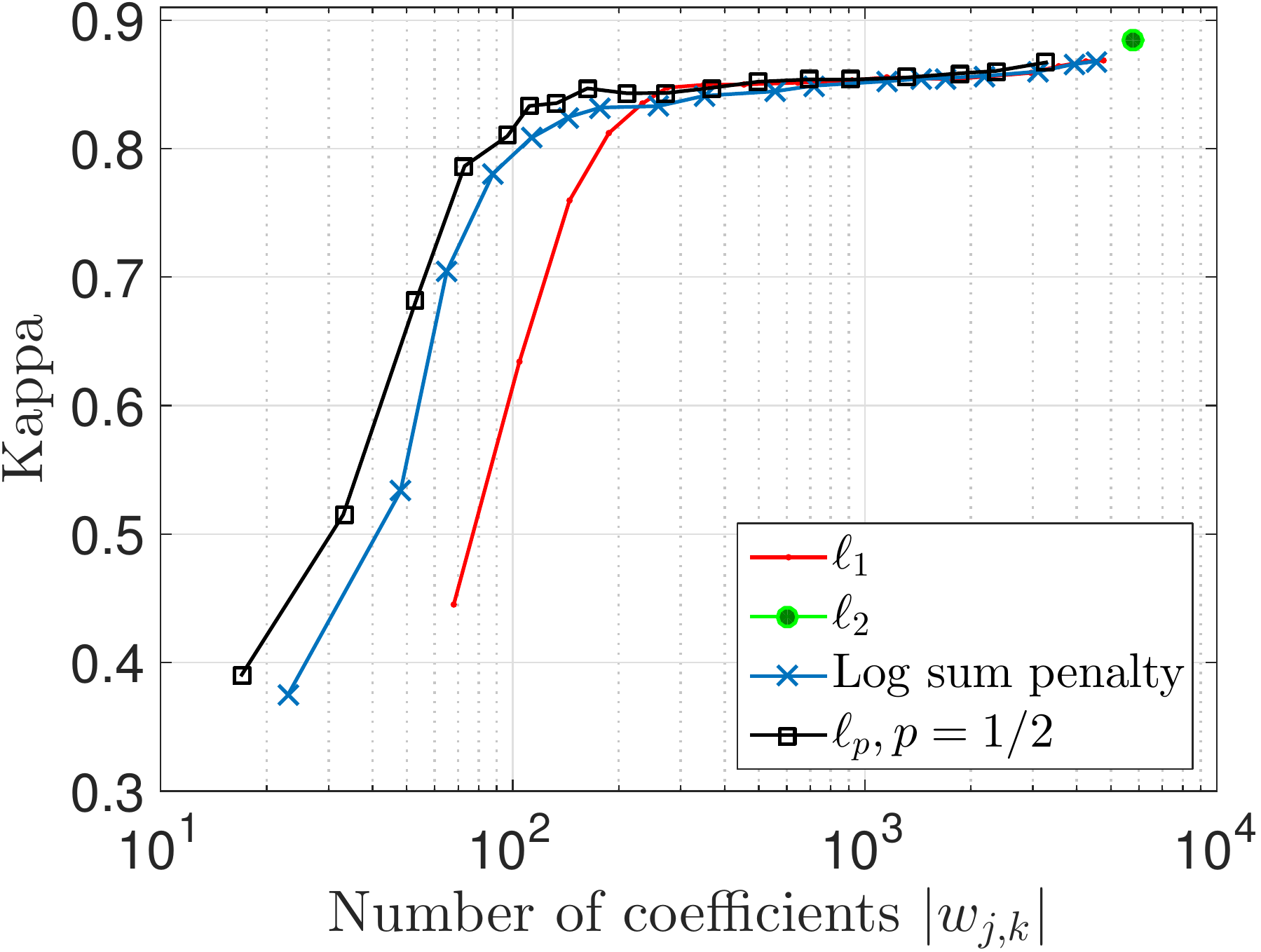} & 
\includegraphics[width=.3\textwidth]{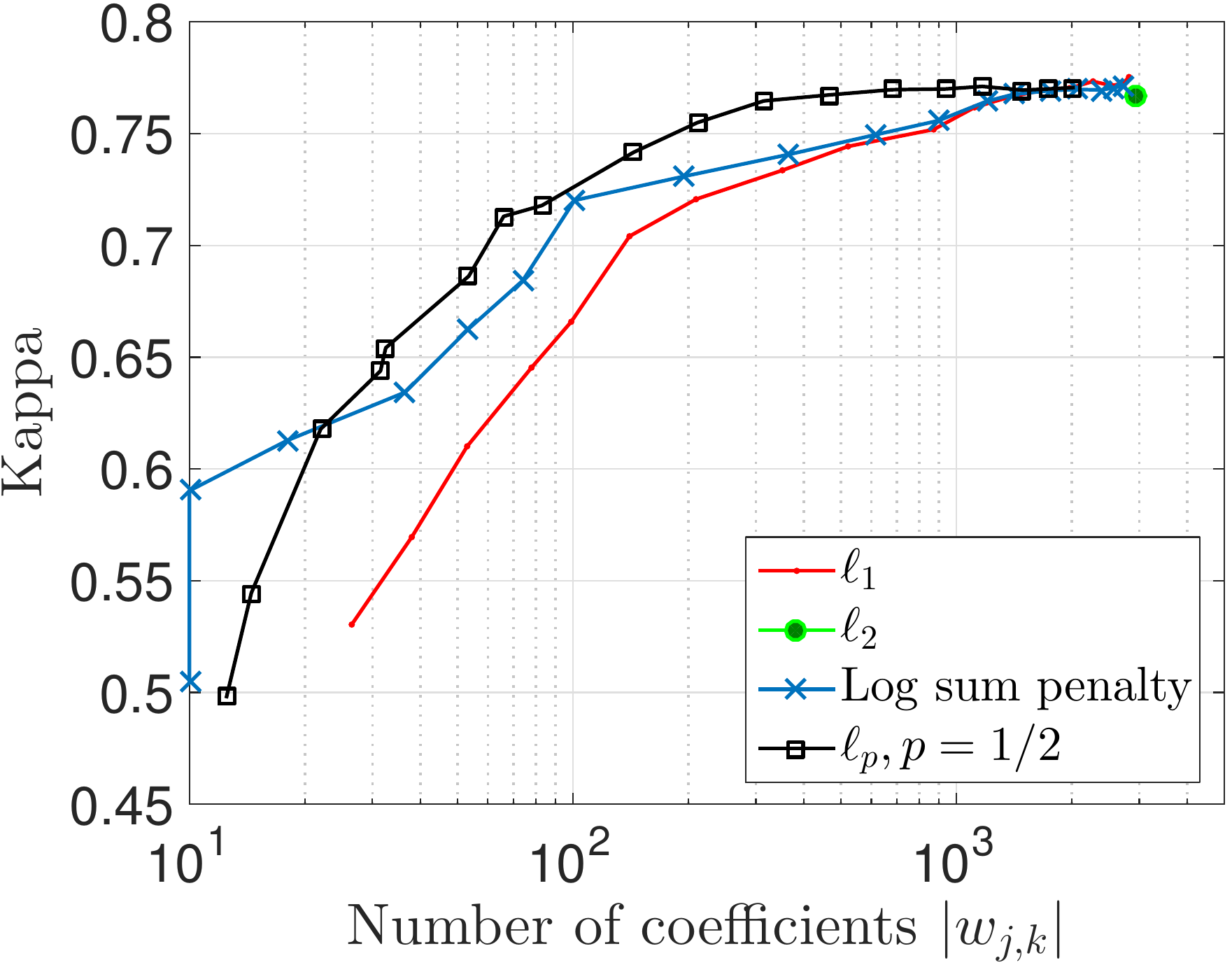} \\
\end{tabular}
\caption{Performance (Kappa) vs. compactness (number of coefficients $w_{j,k} > 0$) for the different regularizers in the \RR{}{\textsc{Thetford mines}, \textsc{Houston} and \textsc{Zurich summer}} datasets.}
\label{fig:res}
\end{figure*}

\noindent{\textbf{Results.}} The results are reported in Fig.~\ref{fig:res}, comparing the regularization paths for the four regularizers and the three datasets presented above. The graphs can be read as a ROC curve: the most desirable situation would be a classifier with both  accuracy and little active features, i.e., a score close to the top-left corner. The $\ell_2$ model shows no variation on the sparsity \RR{axe}{axis} (all the coefficients are active)  and very little variability on the accuracy one: it is therefore represented by a single \green{green} dot. It is remarkably accurate, but is the less compact
model, since it has all the coefficients active. Employing the $\ell_1$ regularizer (\red{red} line), as it is mainly done in the literature of sparse classification, achieves a sharp decrease in the number of
active coefficients, but at the price of a steep decrease in performances of
the classifier. When using 100 active coefficients, the $\ell_1$ model
suffers of a 20\% drop in performance, a trend is observed in all the
experiments reported. 

Using the non-convex regularizers provides the best of both
worlds: the $\ell_p$ regularizer (black line with `$\square$' markers) in particular, but also  the Log sum penalty regularizer (\blue{blue} line with `$\times$' markers) achieve
improvements of about 15-20\% with respect to the $\ell_1$ model. More stable results along the
regularization path are observed: the non-convex regularizers are  less
biased than the $\ell_1$ norm in classification and  achieve
competitive performances with respect to the (non-sparse) $\ell_2$
model with a fraction of the features (around 1-2\%). Note
  that the models of all experiments were initialized with the
  $\mathbf{0}$ vector.\RR{}{ This is sensible for the non-convex
    problem, since all the regularization discussed in the paper (even
    $\ell_2$) tend to shrink
  the model toward this point. 
By initializing at $\mathbf{0}$
 for non-convex regularization, we simply promote a local solution not
too far from this neutral point. In other words one can see the
initialization as an additional regularization.
Moreover the experiments show that the non-convexity leads to
state-of-the-art performance. }

\section{Sparse linear unmixing}
\label{sec:sparse-line-unmix}

In this section we express the sparse linear unmixing problem in the same
optimization framework as Eq.~\eqref{eq:optprob}. We discuss the advantage of
using non-convex optimization. The performance of the $\ell_2$, $\ell_1$ and the non convex $\ell_p$ and LSP 
regularization terms are then compared on a simulated example using
real reflectance spectra (as in \cite{Ior11}). 

\subsection{Model}
\label{sec:model}
Sparse linear unmixing can be expressed as the following optimization problem
\begin{equation}
  \label{eq:linear_unmix}
  \min_{\balpha\geq 0}\quad\frac{1}{2}\|\y-\D\balpha\|_2^2+\lambda R(\balpha),
\end{equation}
where $\y$ is a noisy spectrum observed and $\D$ is a matrix
containing a dictionary of spectra (typically a spectral library).
This formulation adds a positivity constraint to the vector
$\balpha$ \emph{w.r.t.} problem \eqref{eq:optprob}. In practice, \eqref{eq:linear_unmix} can be reformulated as the following unconstrained
optimization problem
 \begin{equation}
  \label{eq:linear_unmix_full}
  \min_{\balpha}\quad\frac{1}{2}\|\y-\D\balpha\|_2^2+\lambda
  R(\balpha)+ \imath_{\balpha\geq 0},
\end{equation}
where $\imath_{\balpha\geq 0}$ is the indicator function that has
value $+\infty$ when one of the component of $\balpha$ is $>0$ and
value $0$ when it is in the positive orthant. By supposing that $\imath_{\balpha\geq 0}$ is
equivalent to  $\lambda\imath_{\balpha\geq 0}, \forall \lambda>0$, we can gather the last two terms into $\tilde R(\balpha)= R(\balpha)+ \imath_{\balpha\geq 0}$, thus leading to a problem similar to Eq.~\eqref{eq:optprob}. All the optimization procedures discussed above
can therefore be used for this reformulation, as long as the proximal operator
\emph{w.r.t.} $\tilde R(\cdot)$ can be computed efficiently. The proximal operator for all
the regularization terms in Table \ref{tab:regterm} with additional positivity
constraints can be obtained by an orthogonal projection on the
positive orthant followed by the proximal of $R$ :
\begin{equation}
  \label{eq:prox_rtilde}
 \text{prox}_{\lambda R+ \imath_{\balpha\geq 0}}(\v)= \text{prox}_{\lambda R}(\max(\v,0)),
\end{equation}
where $\max(\v,0)$ is taken component-wise. This shows that we can use
the exact same algorithm as in the classification experiments of Section~\ref{sec:appl-remote-sens}, since we have
an efficient proximal operator. 

We know that when the solution of Eq.~\eqref{eq:linear_unmix} the resulting 
$\balpha$ must only have a few nonzero components: one might want to promote
more sparsity with a non-differentiable regularization term. Therefore, in the
following we investigate the use of non-convex regularization for linear
unmixing. We focus on problem \eqref{eq:linear_unmix}, but
a large part of the unmixing literature works with an
additional constraint of sum to $1$ for the $\balpha$ coefficients. This
additional prior can sometimes reflect a physical measure and adds
some information to the optimization problem. In our framework, this
constraint can make the direct computation
of the proximal operator non-trivial\RR{ to compute directly}{}. In this case
it is more interesting to use multiple splitting instead of one and to
use other algorithms such as generalized FBS
\cite{raguet2013generalized} or  ADMM, that has already been used for
remote sensing applications \cite{iordache2012total}.

\begin{figure*}
\begin{tabular}{ccc}
\includegraphics[width=.3\linewidth]{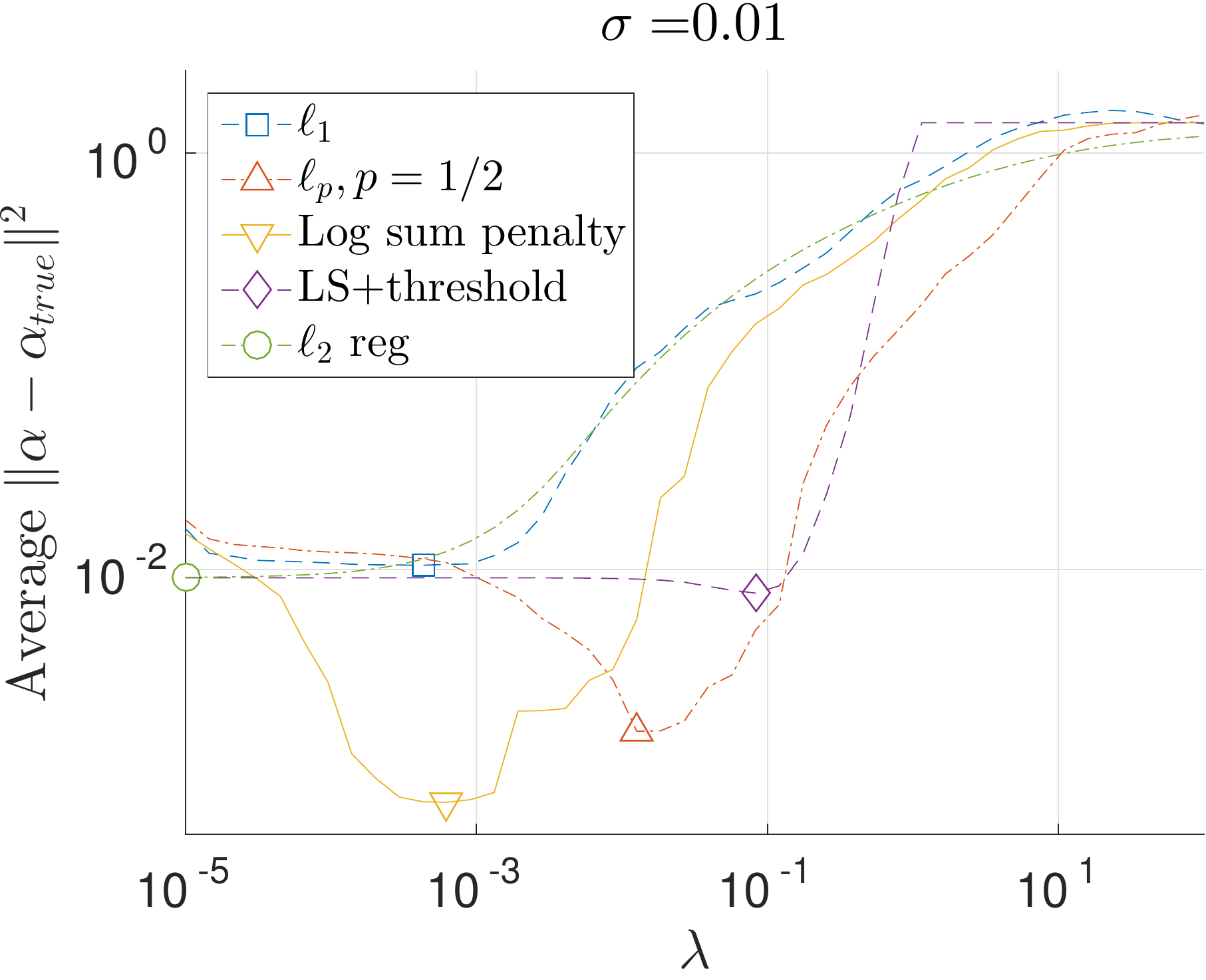}&
\includegraphics[width=.3\linewidth]{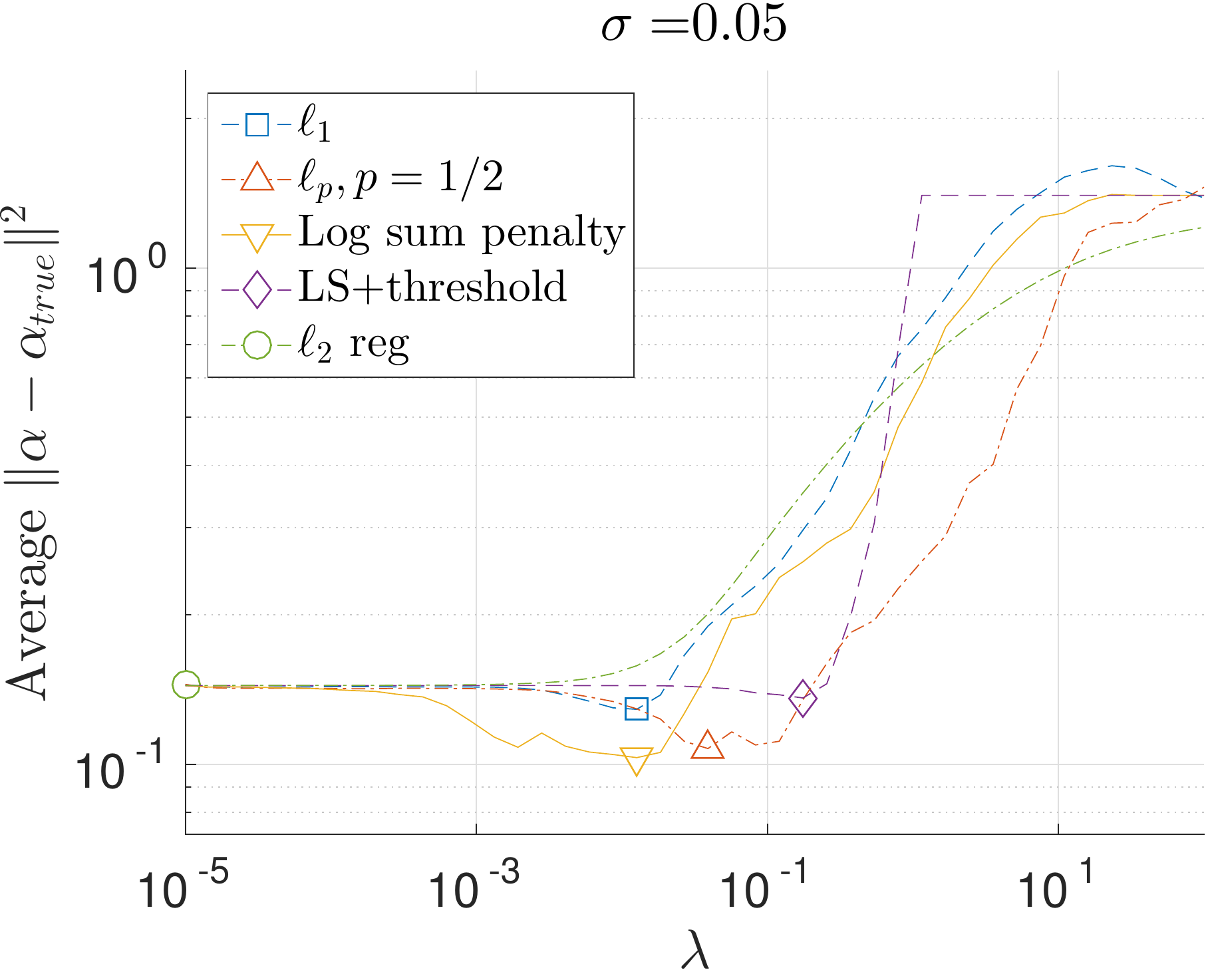}&
\includegraphics[width=.3\linewidth]{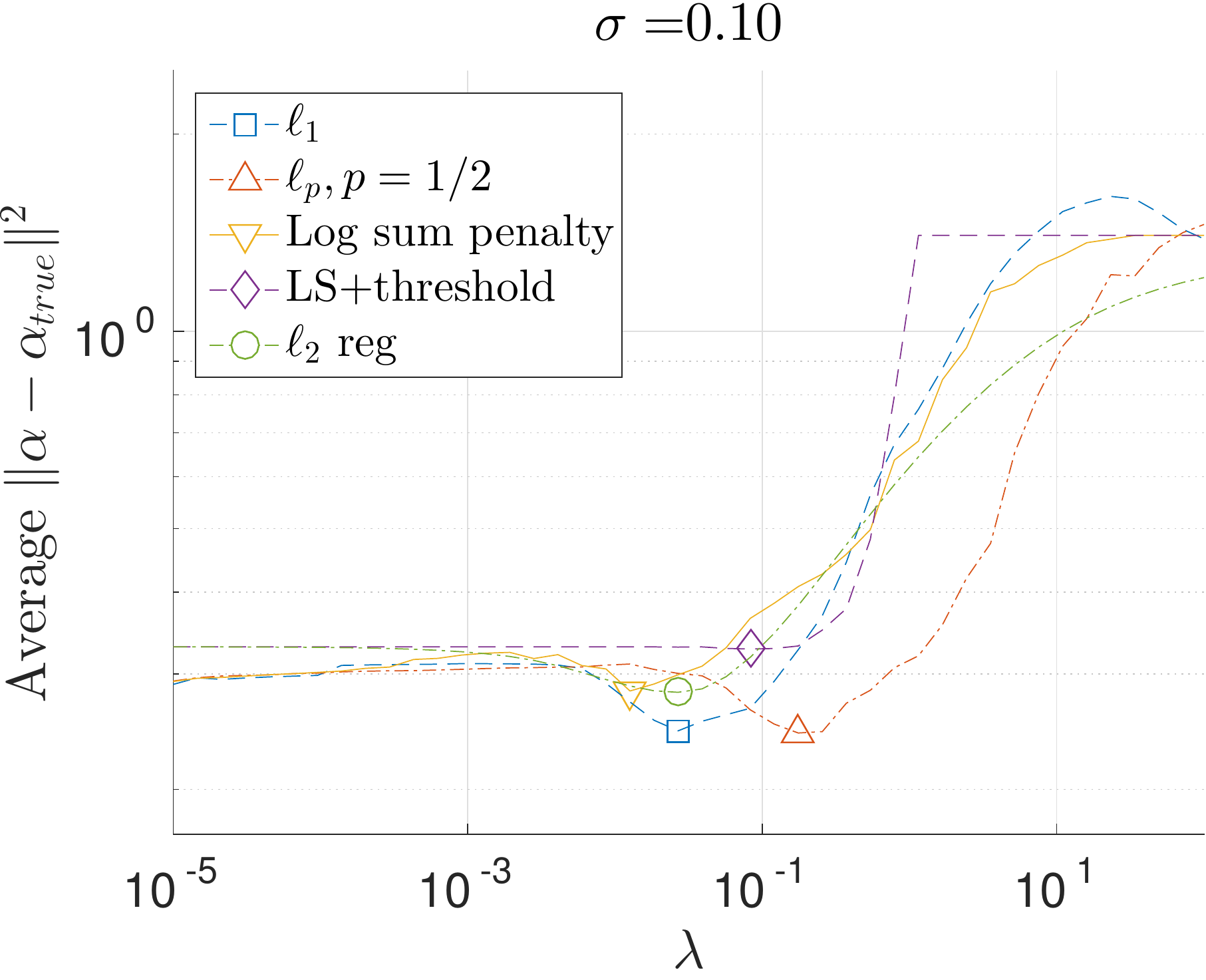}\\
\includegraphics[width=.3\linewidth]{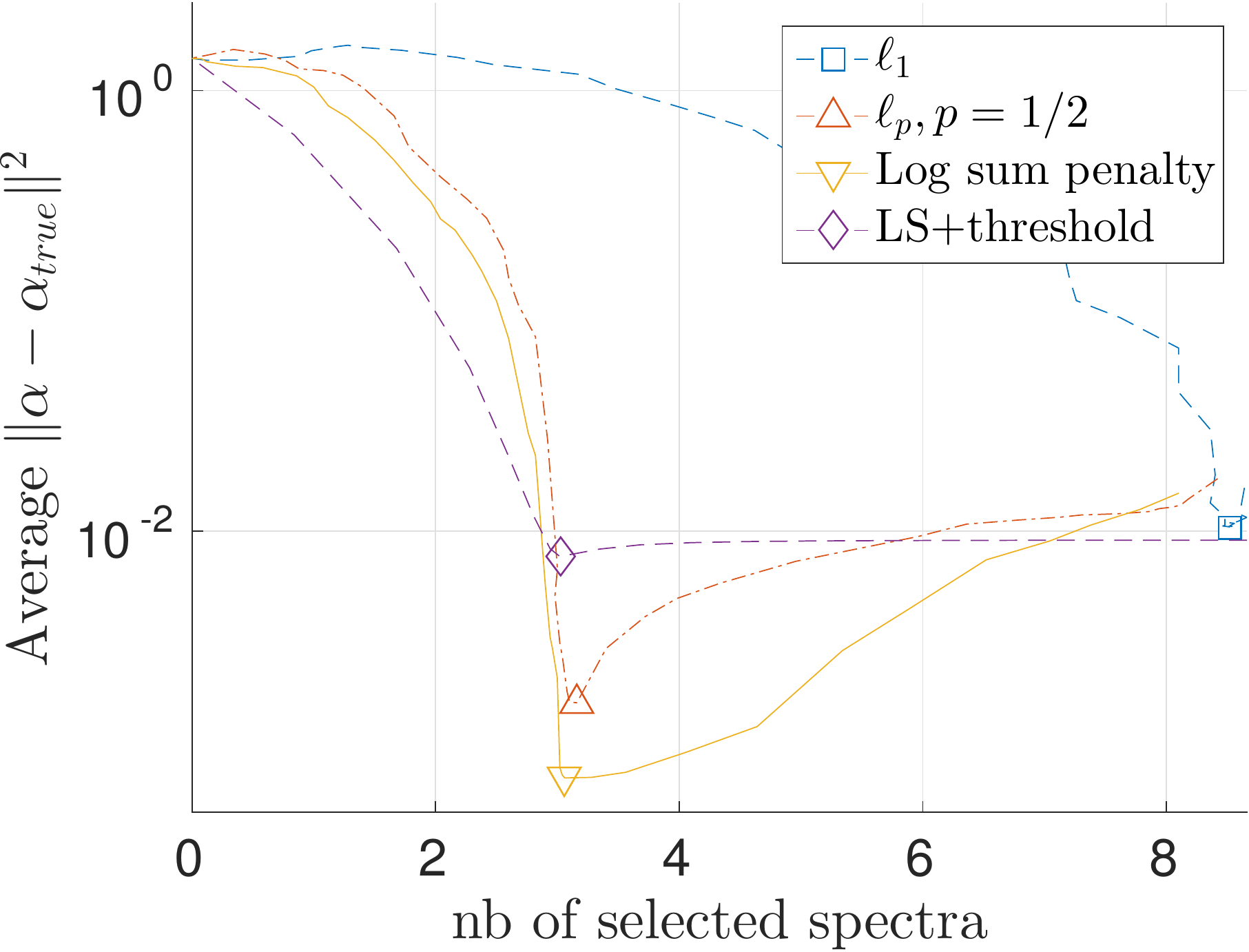}&
\includegraphics[width=.3\linewidth]{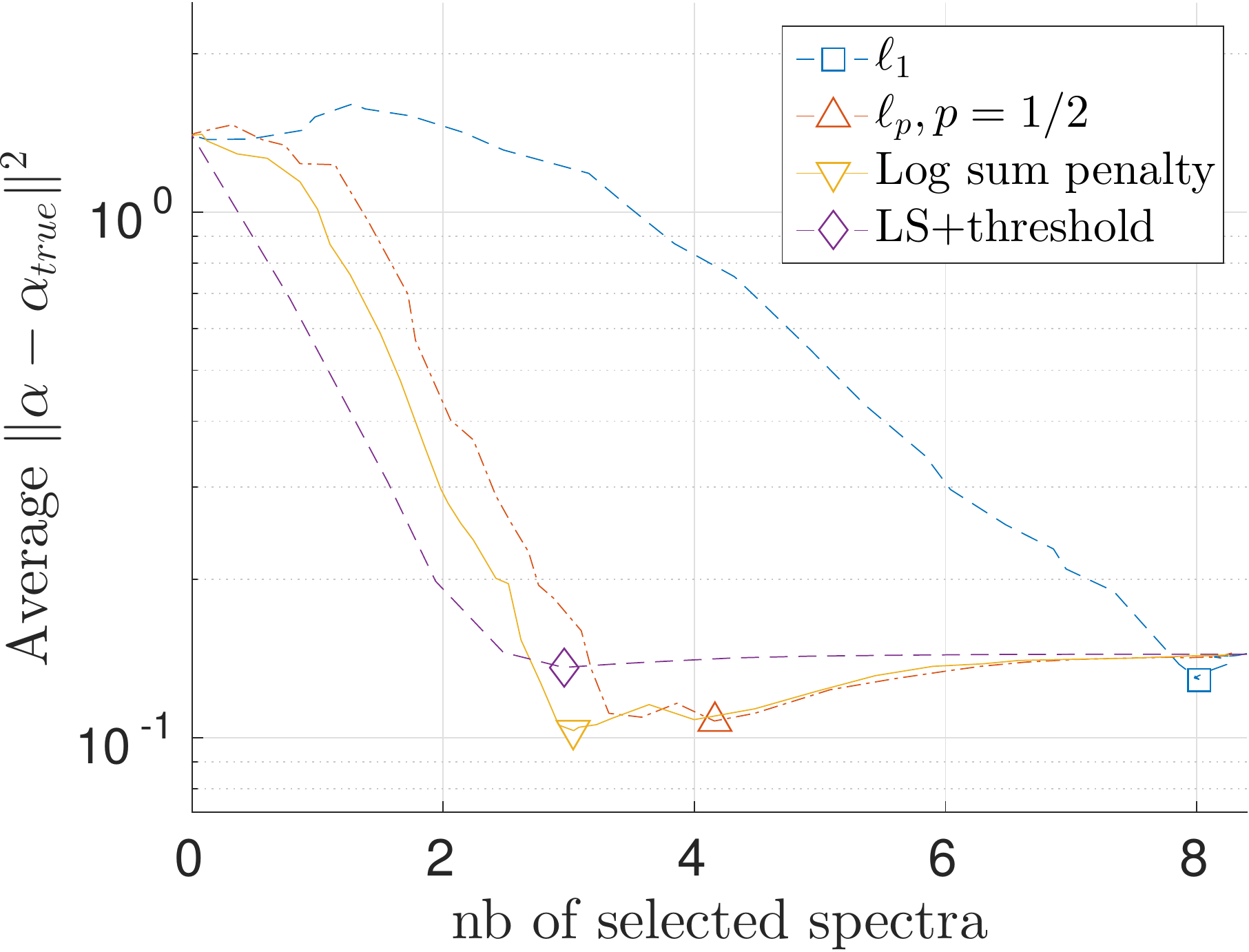}&
\includegraphics[width=.3\linewidth]{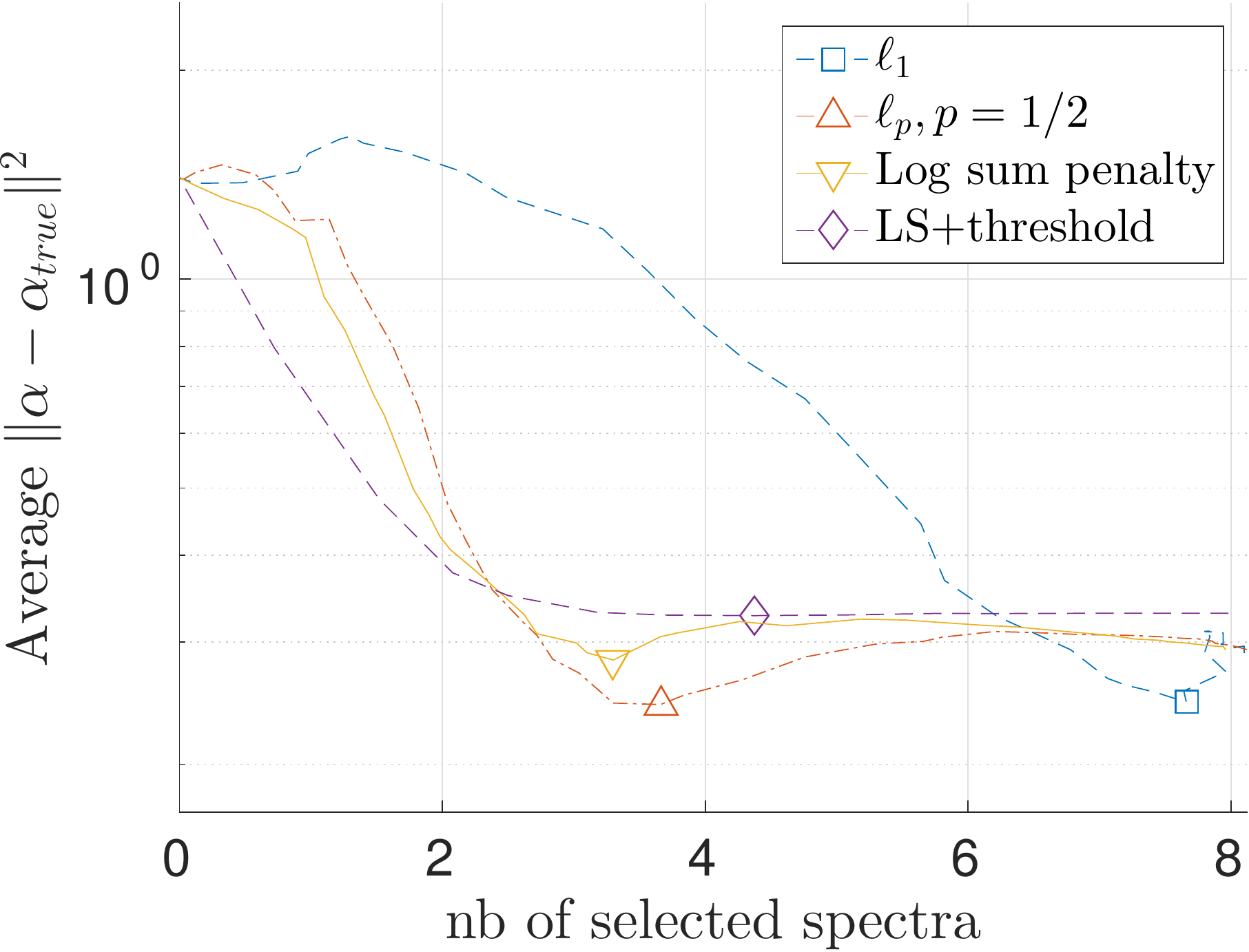}\\
(a)&(b)&(c)
\end{tabular}
\caption{\RR{}{Linear unmixing results on the simulated hyperspectral
  dataset. Each column represetns a different noise level: (a) $\sigma=0.01$ (b)
  $\sigma=0.05$ and (c) $\sigma=0.10$. Model error
  $\|\balpha-\balpha_{true}\|^2$ is plotted either   as a function of the regularization parameter $\lambda$ (top row) of of the number of active coefficients of the final solution (bottom row). The marker show the best  performances of each regularization strategy}.
\label{fig:resunmix}}
\end{figure*}

\subsection{Numerical experiments}
\label{sec:toy-example}
In the unmixing application we consider an example simulated using the
USGS spectral library\footnote{The dataset can be downloaded from
  \url{http://www.lx.it.pt/~bioucas/}}: from the library, we extract
23 spectra corresponding to different materials (by keeping spectra
with less than $15^\circ$ angular distance to each other). \RR{}{Using these
23 base spectra, we simulate mixed  pixels by creating random linear
combinations of $n_{act}\leq 23$ endmembers. The random weight of the
active components are obtained using an uniform random generation in
$[0,1]$ (leading to weights that do not sum to $1$).
We then add to the resulting
signatures some Gaussian noise $n \sim \mathcal{N}(0,\sigma^2)$.
For each numerical experiments we solve the unmixing problem
by least squares with the four regularizers of
Table~\ref{tab:regterm}: $\ell_2$, $\ell_1$, $\ell_p$ and LSP. An
additional approach that consists in performing a hard thresholding on
the positive least square solution (so, the $\ell_2$) has also been investigated (named `LS+threshold' hereafter).
As for
the previous example on classification, we calculate the unmixing
performance on a regularization path, i.e. a series of values of the
regularization parameter $\lambda$ in Eq.~\eqref{eq:linear_unmix},
with $\lambda = [10^{-5}, ..., 10^3]$. We assess the success of the
unmixing by the model error  $\|\balpha-\balpha_{true}\|^2$. We repeat
the simulation 50 times, to account for different combination of the original
elements of the dictionary: all  results reported are averages over
those 50 simulations.}

\RR{}{ First, we compare the different regularization schemes for
  different noise levels (Figure~\ref{fig:resunmix}). We
  set $n_{act}=3$ and report the model error along the regularization
  path (varying $\lambda$) on the top row of
  Figure~\ref{fig:resunmix}. On the bottom
  row, we report the model error as a function of the number of
  selected components, again along the same  regularization path.
We observe that the nonconvex strategies achieve the lowest errors
(triangle shaped markers) on low and medium noise levels, but also that $\ell_p$
seems to be more robust to noise. The $\ell_1$ norm also
achieves good results, in particular in high noise situations. Regarding the error achieved
per level of sparsity
(represented in the bottom row of Fig.~\ref{fig:resunmix}) , we observe that
the nonconvex regularizers achieve
far better reconstruction errors, in particular around the right
number of active coefficient (here $n_{act}=3$).  On average, the best
results are obtained by the the LSP and
$\ell_p$ regularization. Note that the $\ell_1$ regularizer needs a larger number
of active component in order to achieve good model reconstruction (of
the order of $9$ when the actual number of coefficient is $3$). The
LS+threshold approach seem to work well for component selection, but
leads to an important decrease in accuracy of the model.}

\RR{}{In order to evaluate the ability of a method to estimate a good
  model and select the good active components at the same time, we run
simulations with a fixed noise level $\sigma=0.05$ but for a varying
number of true active components $n_{act}$, from $1$ to $23$. In this
configuration, we first find for all
regularizations the smallest $\lambda$ that leads to the correct
number of selected component $n_{sel}=n_{act}$. The average model
error as a function of $n_{act}$ is reported in Figure
\ref{fig:resunmix_nactive}(a). We can see that the non-convex
regularization leads to better performances when the correct number
of spectra is selected (compared to $\ell_1$ and LS+threshold). In Figure \ref{fig:resunmix_nactive}(b) we report the number of selected components
as a function of the true number of active components when the model
error is minimal. We observe that nonconvex regularization manages to both
select the correct components and estimate a good model when a small
number of components are active ($n_{act} \leq 10$), but also that it fails (as $\ell_1$ does) for large
numbers of active components. This result illustrates the fact that
non-convex regularization is more aggressive in term of sparsity and obviously
performs best when sparsity is truly needed. }

\begin{figure}
\begin{tabular}{cc}
\includegraphics[width=.45\linewidth]{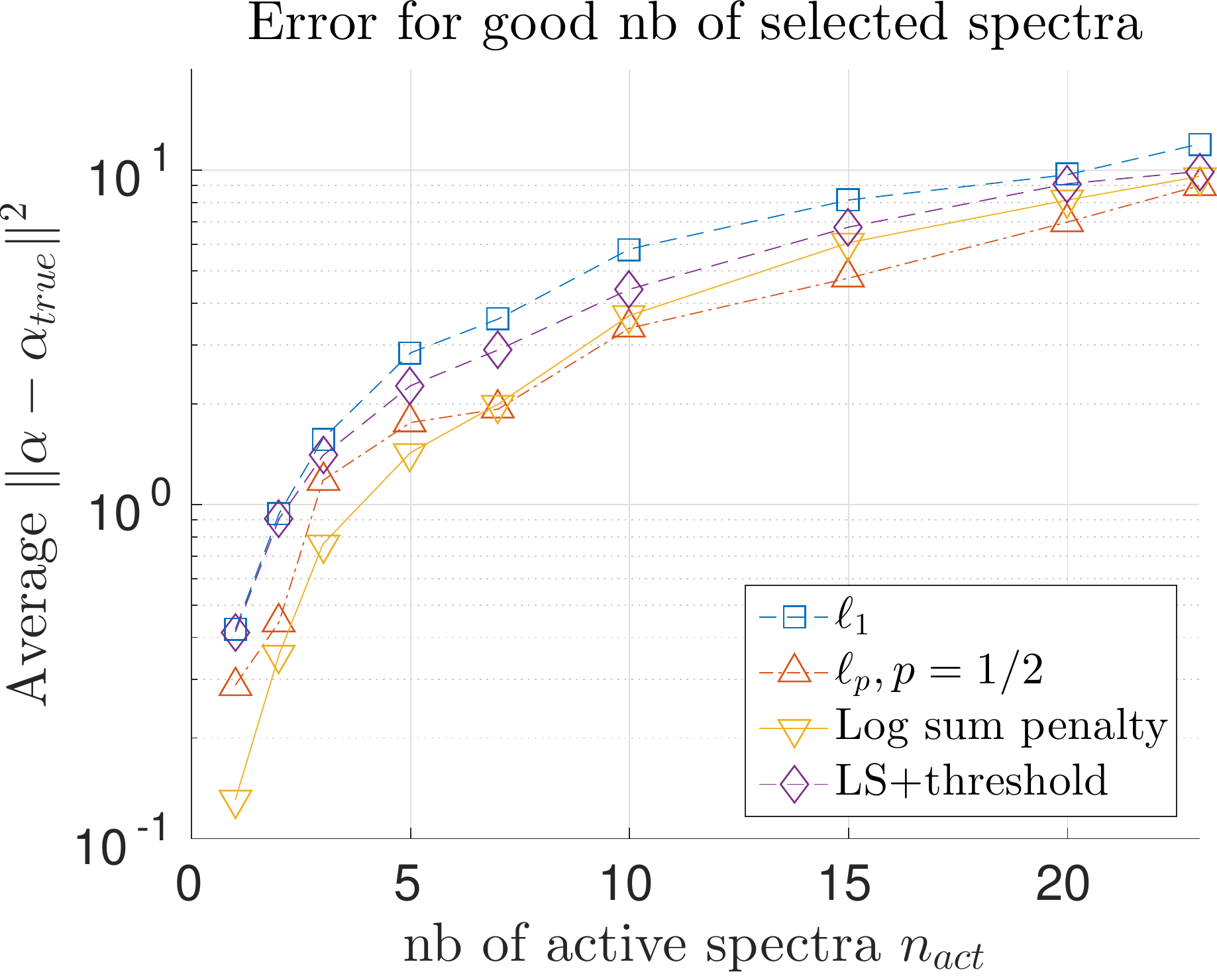}&
\includegraphics[width=.45\linewidth]{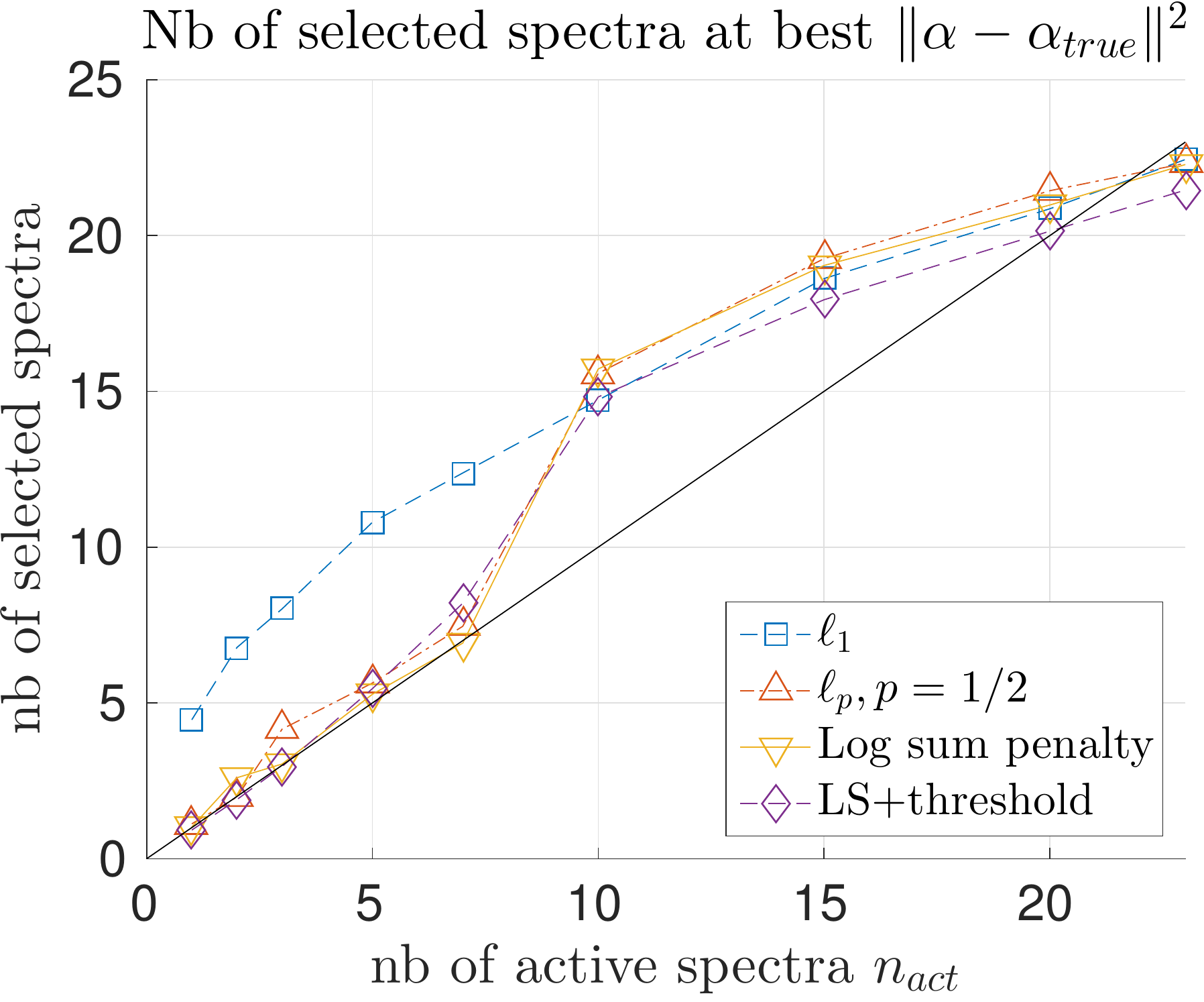}\\
(a) &(b)
\end{tabular}
\caption{\RR{}{Linear unmixing results on the simulated hyperspectral
  dataset for increasing number of active spectra in the mixture: (a) model error, for the best solution with the number of selected spectra closest to $n_{act}$ and (b) number of selected spectra for the model with the lowest error.} 
  \label{fig:resunmix_nactive}}
\end{figure}

\section{Conclusions}
\label{sec:conclusions}
In this paper, we presented a general framework for non-convex regularization
in remote sensing image processing. We discussed different ways to
promote sparsity and avoid the bias when sparsity is required  
via the use of non-convex regularizers. We applied the proposed
regularization schemes to problems of hyperspectral image
classification and linear unmixing:  in all scenarios, we showed that
non-convex regularization leads to the best performances when
accounting for both sparsity and quality of the final product. Non
convex regularizers promote compact solutions, but without the bias
(and the decrease in performance) related to nondifferentiable convex norms such as the popular $\ell_1$
norm. \\ Non convex regularization is a flexible and general framework
that can be applied to every regularized processing scheme: keeping
this in mind, we also provide a toolbox to the community to apply
non-convex regularization to a wider number of problems. \RR{}{The
  toolbox can now be accessed online (see also the Appendix of this article for a description of the toolbox).}}

\section{Acknowledgements}
The authors would like to thank Telops Inc. (Qu\'ebec, Canada) for acquiring and providing the \textsc{Thetford mines} data, as well as the IEEE GRSS Image Analysis and Data Fusion Technical Committee (IADFTC) and Dr. Michal Shimoni (Signal and Image Centre, Royal Military Academy, Belgium) for organizing the 2014 Data Fusion Contest, the Centre de Recherche Public Gabriel Lippmann (CRPGL, Luxembourg) and Dr. Martin Schlerf (CRPGL) for their contribution of the Hyper-Cam LWIR sensor, and Dr. Michaela De Martino (University of Genoa, Italy) for her contribution to data preparation. \\
The authors would like to thank the Hyperspectral Image Analysis group and the NSF Funded Center for Airborne Laser Mapping (NCALM) at the University of Houston for providing the \textsc{Houston} data sets and the IEEE GRSS IADFTC for organizing the 2013 Data Fusion Contest.\\
The authors would like to acknowledge Dr. M. Volpi and Dr. Longbotham for making ther \textsc{Zurich Summer} data available. \\
The authors would like to acknowledge Dr. Iordache and Dr. Bioucas-Dias for sharing the USGS library used in the unmixing experiment.

\section*{Appendix}
\subsection{\RR{}{Optimization toolbox}}

\RR{}{In order to promote the use of non-convex regularization in the
  remote sensing community, we provide the reader with a simple to use
  Matlab/Octave generic optimization toolbox. The code will provide
  a generic solver (complete rewriting of GIST) for problem
  \eqref{eq:optprob} that is able to
  handle a number of regularization terms (at least all the terms in
  Table \ref{tab:regterm}) and any differentiable data fitting term
  $L$. We provide several function for performing multiclass
  classification tasks such as
  SVM, logistic regression and calibrated hinge loss. For linear
  unmixing we provide the least square loss, but extension to other
  possibly more robust data fitting terms can be performed easily. For
  instance, performing
  unmixing with the more robust Huber loss \cite{huber1964robust}
  would require the change of  only two lines in function \texttt{``gist\_least.m''},
  \emph{i.e.} the computation of the
  Huber loss and its gradient. The toolbox can now be accessed
    at \url{https://github.com/rflamary/nonconvex-optimization}. It
    is freely available on Github.com  as a community
  project and we welcome contributions.}

\bibliographystyle{IEEEbibS}

\end{document}